\documentclass[sigconf]{acmart}

\usepackage{acmart-taps}

\usepackage{tikz}
\usepackage{amsmath}
\usepackage{textcomp}
\usepackage{colortbl}
\usepackage{eucal}
\usepackage{bm}
\usepackage{algorithm}
\usepackage{subfigure}
\usepackage{booktabs}
\usepackage{multirow}
\usepackage{enumitem}
\usepackage{url}
\usepackage{xspace}
\usepackage{amsfonts}
\usepackage{subcaption}
\usepackage{booktabs}
\usepackage{microtype}
\usepackage{textcomp}
\usepackage[normalem]{ulem}

\def\ourSystem{\textit{SoilX}\xspace}
\def\ie{\textit{i.e.},\xspace}
\def\eg{\textit{e.g.},\xspace}
\def\etl{\textit{et al.}\xspace}
\def\vs{\textit{vs.}\xspace}

\newcolumntype{C}[1]{>{\centering\arraybackslash}p{#1}}

\newcolumntype{L}[1]{>{\raggedright\arraybackslash}p{#1}}

\copyrightyear{2026}
\acmYear{2026}
\setcopyright{cc}
\setcctype{by-nc-nd}
\acmConference[SenSys '26]{ACM/IEEE International Conference on Embedded Artificial Intelligence and Sensing Systems}{May 11--14, 2026}{Saint Malo, France}
\acmBooktitle{ACM/IEEE International Conference on Embedded Artificial Intelligence and Sensing Systems (SenSys '26), May 11--14, 2026, Saint Malo, France}
\acmDOI{10.1145/3774906.3800499}
\acmISBN{979-8-4007-2309-4/2026/05}

\begin{document}

\title{\ourSystem: Calibration-Free Comprehensive Soil Sensing through Contrastive Cross-Component Learning}

\author{Kang Yang}
\authornote{The first three authors contributed equally to the work.}
\orcid{0000-0001-8248-4894}
\affiliation{\institution{University of California, Merced}
\city{Merced}
\country{USA}}
\email{kyang73@merced.edu}

\author{Yuanlin Yang}
\authornotemark[1]
\orcid{0009-0002-5670-5985}
\affiliation{\institution{University of California, Merced}
\city{Merced}
\country{USA}}
\email{yuanlinyang@merced.edu}

\author{Yuning Chen}
\authornotemark[1]
\orcid{0009-0002-2814-1166}
\affiliation{\institution{University of California, Merced}
\city{Merced}
\country{USA}}
\email{ychen372@merced.edu}

\author{Sikai Yang}
\orcid{0000-0003-0959-5867}
\affiliation{\institution{University of California, Merced}
\city{Merced}
\country{USA}}
\email{syang126@merced.edu}

\author{Xinyu Zhang}
\orcid{0000-0001-9688-8056}
\affiliation{\institution{University of California, San Diego}
\city{San Diego}
\country{USA}}
\email{xyzhang@ucsd.edu}

\author{Wan Du}
\orcid{0000-0002-2732-6954}
\affiliation{\institution{University of California, Merced}
\city{Merced}
\country{USA}}
\email{wdu3@merced.edu}

\begin{abstract}

Precision agriculture demands continuous and accurate monitoring of soil moisture~$M$ and key macronutrients, including nitrogen~$N$, phosphorus~$P$, and potassium~$K$, to optimize yields and conserve resources.
Wireless soil sensing has been explored to measure these four components; however, current solutions require recalibration~(\ie re-train the data processing model) to handle variations in soil texture~(characterized by aluminosilicates~$Al$ and organic carbon~$C$, limiting their practicality.
To address this, we introduce \ourSystem, a calibration-free soil sensing system that jointly measures six key components:~$\{M, N, P, K, C, Al\}$.
By explicitly modeling~$C$ and~$Al$, \ourSystem eliminates texture- and carbon-dependent recalibration.
\ourSystem incorporates Contrastive Cross-Component Learning~(3CL), with two customized terms: the Orthogonality Regularizer and Separation Loss, to effectively disentangle cross-component interference.
Additionally, we design a novel Tetrahedral Antenna Array with an antenna-switching mechanism, which can robustly measure soil dielectric permittivity independent of device placement.
Extensive experiments demonstrate that \ourSystem reduces estimation errors by 23.8\% to 31.5\% over baselines and generalizes well to unseen fields.

\end{abstract}

\ccsdesc[500]{Computing methodologies}
\ccsdesc[500]{Computing methodologies~Machine learning algorithms}
\ccsdesc[500]{Computer systems organization~Sensor networks}

\keywords{Wireless Sensing, Soil Sensing, Contrastive Learning}

\maketitle

\section{{Introduction}}\label{sec_introduction}

Precision agriculture necessitates accurate and continuous monitoring of soil moisture~$M$ and macronutrients such as nitrogen~$N$, phosphorus~$P$, and potassium~$K$ across farm regions~\cite{wang2024soilcares, smart_argi}. 
It enhances crop yields, optimizes irrigation, and reduces aquatic pollution from fertilizer overuse~\cite{farooq2020role, galal2024soil}.
Recently, wireless systems have been used to measure soil moisture~\cite{ding2019towards, wang2020soil, chang2022sensor, khan2022estimating, feng2022lte, jiao2023soiltag, ding2023soil, ren2024demeter, chen2024metasoil} and macronutrients~\cite{wang2024soilcares}.
SoilCares~\cite{wang2024soilcares} measures four soil components,~\(\{M, N, P, K\}\) by integrating Long Range~(LoRa)~\cite{sensys_lldpc, ipsn_flog} and Visible and Near-Infrared~(VNIR) sensing as complementary modalities.
However, SoilCares requires calibration to address variations in soil texture~$Al$ and organic carbon~$C$, involving retraining with newly collected, soil-specific ground truth data.
According to our experiments in~§\ref{sec_interference}, when testing data has different carbon content from training data, macronutrient estimation error can increase by $1.25\times$.
Calibration is a significant problem in soil sensing as collecting ground truth data~(labels) is costly and time-consuming.
No in-field devices can provide sufficient soil analysis accuracy.
We need to send new soil samples to a soil analysis laboratory and wait weeks for the results.

To this end, this paper presents \ourSystem, a calibration-free soil sensing system that jointly measures six components~\(\{M, N, P, K, C, Al\}\). 
By explicitly modeling~$C$ and~$Al$, \ourSystem eliminates texture- and carbon-dependent recalibration.
For instance, $Al$ differentiates the clay~(high~$Al$ content) from sandy soils~(quartz-dominated, low~$Al$), while~$C$ quantifies organic matter’s impact on porosity and moisture retention~\cite{ben1995near, matichenkov2001relationship}.
As a result, \ourSystem maintains high sensing accuracy on unknown soil samples without retraining.

\ourSystem also builds on LoRa and VNIR sensing, like SoilCares~\cite{wang2024soilcares}.
However, \ourSystem makes distinct contributions by tackling two key challenges.
\text{i)}~Jointly measuring~\(\{M, N, P, K, C, Al\}\) is challenging due to cross-component interference, \ie bidirectional interactions among soil components collectively distort~LoRa and~VNIR signal responses~\cite{rossel2006visible, soriano2014performance}.
For instance, high moisture~\(M\) in clay-rich soils mimics the dielectric permittivity of sand, distorting LoRa signals for carbon~\(C\) and~$NPK$.  
Conversely, carbon~\(C\) enhances water retention by increasing porosity, further impacting~\(M\) signal responses~\cite{ben1995near, rawls2003effect}.
Similarly, carbon~\(C\) scatters VNIR light, blending the spectral signatures of~$NPK$, while~\(P\) alters soil pH, shifting~VNIR reflectance and affecting~\(C\) detection~\cite{janzen2006soil, doolittle2014use}.
\text{ii)}~Existing LoRa dual-antenna designs for measuring soil dielectric permittivity~$\epsilon$ impose strict requirements on antenna placement~\cite{chang2022sensor, wang2024soilcares}.
Specifically, the line connecting the two antennas needs to be oriented perpendicular to the soil surface~\cite{chang2022sensor, wang2024soilcares}, limiting their practicality.

\textbf{i)~For the cross-component interference challenge,} Contrastive Learning~(CL)~\cite{radford2021learning} offers a promising solution by learning latent representations to capture subtle variations in soil components.
In particular,~CL compares samples by forming positive pairs with their augmentations and negative pairs with others, minimizing distances for positive pairs and maximizing them for negative ones.
This enables~CL to disentangle bidirectional interference by contrasting various soil samples.
Despite their promise, existing~CL methods are ill-suited for the regression task of soil sensing due to three limitations.
\textit{First,} they are designed for classification with contrastive losses like InfoNCE~\cite{rusak2024infonce}, where binary positive-negative pairing disrupts the ordinal nature of soil components.  
\textit{Second,} regression adaptations such as RnC~\cite{zha2023rank} preserve order in single-output tasks but fail to address multi-output interference.
\textit{Third,}~CL's reliance on large datasets is impractical due to the high cost and labor-intensive nature of soil data collection~\cite{yang2025gwrf, yang2025gsrf}.

To address these limitations, \ourSystem introduces a novel~Contrastive Cross-Component Learning~(3CL) framework that integrates a wireless channel encoder and a compound contrastive loss, combining an Orthogonality Regularizer and a Separation Loss to disentangle cross-component interference and accurately measure six soil components~\(\{M, N, P, K, C, Al\}\) using limited training data.

Specifically, each soil sample is characterized by a sensing vector, formed by concatenating soil permittivity from LoRa and VNIR spectroscopy.  
The wireless channel encoder maps each vector into a~$D$-dimensional latent space, where the influence of each soil component is implicitly embedded as a directional vector.
However, these influences may overlap, causing cross-component interference.
To disentangle them, we orthogonalize each component’s directional vector.  
We first customize a training dataset, which can be partitioned into six groups, each containing a limited number of samples where only one component's value varies while the others remain fixed.
Comparing latent embeddings within each group yields a direction vector representing the influence of that component, as only its values vary.
The Orthogonality Regularizer then enforces mutual orthogonality among these vectors, decoupling cross-component interference.
Additionally, at the sample level, we impose that distances between sample embeddings align with differences in their ground truth component values, achieved through the Separation Loss.
Together, the Orthogonality Regularizer and Separation Loss form a compound contrastive loss that preserves relative distances while disentangling component effects.

\ourSystem's 3CL framework adopts a pre-training scheme~\cite{radford2021learning}, leveraging~3CL to learn efficient representations.
Once pre-trained, \ourSystem requires no additional fine-tuning or re-training during inference.

\textbf{ii)~For the strict antenna placement challenge,} \ourSystem introduces the Easy-to-Use~(E2U) Tetrahedron Antenna Array.  
Four antennas are positioned at the vertices of an equal-length tetrahedron and connected to the LoRa Radio via a Single Pole Four Throw~(SPFT) switch.  
During each LoRa packet’s preamble transmission, the SPFT switch sequentially activates each antenna.
First, switching between any two antennas cancels out phase measurement errors caused by Carrier Frequency Offset~(CFO) and baseband signal variations~\cite{wang2024soilcares}.
Second, the tetrahedron-based antenna array leverages its 3D geometry to compute soil permittivity~$\epsilon$ from phase differences across multiple spatial paths.
This spatial diversity eliminates strict placement constraints, allowing LoRa nodes to be freely buried in the soil regardless of antenna orientation, thereby improving device deployment flexibility in real-world scenarios.

For the VNIR device, the transceivers are embedded in a Printed Circuit Board~(PCB), with a photodiode at the center serving as the receiver, surrounded by seven Light Emitting Diodes~(LEDs)~\cite{Bulb}.
The emitted light is absorbed and reflected by a membrane~\cite{Membrane} covered by soil, then received by the photodiode. 
Thus, VNIR does not suffer from strict placement constraints, as evidenced in §\ref{sec_vnir_plane}.

We evaluate \ourSystem by controlled in-lab experiments and in-field tests.
In-lab experiments under varying soil conditions and device placements show that \ourSystem achieves average estimation errors of 4.17\% for~$M$, 4.19\% for~$C$, 1.49\% for~$Al$, and 2.61\textperthousand, 3.56\textperthousand, and 2.79\textperthousand{} for~$N$,~$P$, and~$K$, respectively.
This reflects a performance gain of 23.8 to 31.5\% over the state-of-the-art method, SoilCares~\cite{wang2024soilcares}, in measuring four soil components~\(\{M, N, P, K\}\).
In-field experiments conducted over 12 days across five distinct soil types under diverse weather conditions further validate \ourSystem's effectiveness in accurate multi-component sensing in real-world environments.

In summary, this paper makes the following key contributions:
\begin{itemize}[label=\textbullet, leftmargin=1em]

\item We present \ourSystem, a calibration-free soil sensing platform that integrates LoRa-based permittivity and VNIR spectroscopy for simultaneous multi-component estimation.

\item We develop the 3CL framework to suppress cross-component interference. 
Compared with traditional contrastive learning, 3CL enables modeling of continuous multi-variable correlations.

\item We design a regular tetrahedral LoRa antenna array that enables orientation-invariant permittivity measurement.

\item We demonstrate \ourSystem through both laboratory and field experiments under diverse soil conditions, confirming its effectiveness.

Source code of \ourSystem is available at \textbf{\url{https://github.com/ycucm/soilx_3CL}}.

\end{itemize}

\section{{Background and Motivation}}

We begin by introducing the importance of measuring six soil components, and then demonstrate the limitations of existing systems.

\subsection{A Primer on Soil Sensing}

\begin{figure}
\centering
{\includegraphics[width=0.45\textwidth]{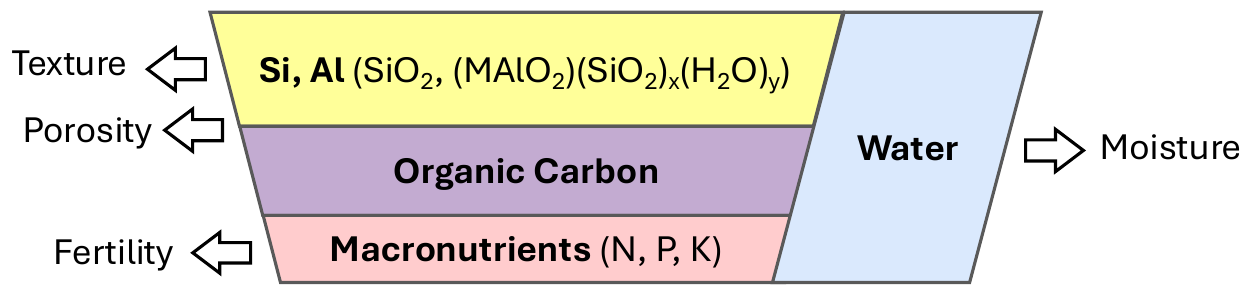}}

\caption{Overview of several key soil components and their effect on soil properties.}
\Description[]{}
    \label{fig_tri}
 
\end{figure}

Figure~\ref{fig_tri} illustrates major soil components, including aluminosilicates, organic carbon, water, and macronutrients, and their effects on porosity, texture, and fertility.  
Aluminosilicates are expressed as~$\text{X}_l \text{Al}_m \text{Si}_n \text{O}_w \cdot n\text{H}_2\text{O}$, where $\text{X}$ denotes metal cations.  
Organic carbon enhances microbial activity~\cite{paul2023soil}, and macronutrients~$NPK$ support crop productivity~\cite{clarkson1980mineral}.  
Together, these components shape soil fertility, structure, and water dynamics.

\textbf{Accuracy Requirement.}
Accurate soil assessment supports decisions on crop selection, irrigation, and fertilization~\cite{Soiltexture, singh2011soil, lal2004soil, brady2008nature}.
For example, $M$ regulates irrigation, where a 1–2\% error may affect efficiency in water-stressed regions~\cite{soulis2015investigating}.  
$NPK$ influence nutrient availability, with a 2\% error in $N$ potentially leading to over- or under-fertilization~\cite{sitters2020nutrient}. 
$C$ improves soil structure and carbon storage, tolerating 3–4\% error~\cite{lark2015implicit}.  
$Al$ reflects texture variation~(\eg 39.5\% in clay \vs 1\% in sand)~\cite{foth1978fundamentals}, making it more error-tolerant.

We exclude silicon~$Si$ because measuring $\{M,N,P,K,C,Al\}$ suffices to guide irrigation and fertility management, and $Si$ can be inferred as a residual from these components. 
Specifically, in soils, $Si$ is bound within aluminosilicate minerals and varies only on geologic time scales, while its plant-available form is largely governed by texture and moisture~\cite{coskun2019controversies}.

\textbf{Temporal Dynamics.}
Soil components differ markedly in their temporal behaviors and underlying drivers.
Moisture~$M$ varies within minutes to hours due to rainfall, irrigation, and evapotranspiration, exhibiting strong short-term dynamics.
Macronutrients~$NPK$ evolve over days to weeks, driven by fertilization, plant uptake, microbial activity, and leaching.
Organic carbon~$C$ changes seasonally or over multiple years as litter decomposes and microbes process organic matter.
Aluminosilicate texture~$Al$ remains stable on agronomic time scales, representing mineral composition and serving as a structural baseline.

Conventional in-situ probes~(\eg capacitance or TDR meters) measure~$M$ reliably but require frequent calibration and laboratory analysis.
LoRa-derived permittivity is sensitive to~$M$ and complementary to~$C$ and~$Al$, while VNIR spectroscopy provides diagnostic information for $NPK$ and~$C$. 
By integrating these modalities, \ourSystem captures both fast- and slow-varying soil properties with a single node, eliminating the need for per-site recalibration.

\subsection{RF-VNIR-Based Soil Sensing}

To measure multiple soil components, existing works integrate~radio frequency~(RF) and VNIR as complementary sensing modalities~\cite{wang2024soilcares, ding2024cost}.
RF sensing measures~$M$ by analyzing soil dielectric permittivity~$\epsilon$ but is prone to interference from uneven surfaces, rocks, and roots.
Additionally, it is insensitive to chemical composition, making it unsuitable for detecting macronutrient.  
Conversely, VNIR sensing excels at detecting~$\{N, P, K, C, Al\}$ via analyzing spectral absorption but is affected by moisture-induced variations, soil texture, and particle size~\cite{shao2011nitrogen}.

For example, SoilCares~\cite{wang2024soilcares} follows a sequential approach: it first estimates moisture~$M$ using LoRa-derived permittivity~$\epsilon$ measurements and VNIR reflectance via Random Forest regression model, then corrects VNIR spectral measurements for moisture interference to estimate~$NPK$.
Similarly, Scarf~\cite{ding2024cost} combines Wi-Fi signals for moisture estimation with image-based lightness correction to estimate organic carbon~$C$ content.

In this work, among RF technologies, LoRa is selected for soil permittivity~$\epsilon$ measurement due to its long-range and low-power communication capabilities.
Unlike Wi-Fi or RFID, which have limited sensing ranges~\cite{ding2019towards, jiao2023soiltag}, LoRa provides broader coverage~\cite{hou2025MoLoRa, ruonan2025Interference, xia2023pcube} and deeper soil penetration, as validated in our experiments~(§\ref{sec_exp_depth} and~§\ref{sec_sensing_range}), making it well-suited for agricultural applications.

\subsection{Calibration Requirement}\label{sec_interference}

Recent soil sensing systems provide multi-component measurements, \ie SoilCares~\cite{wang2024soilcares} for~\(\{M, N, P, K\}\) and Scarf~\cite{ding2024cost} for~\(\{M, C\}\).  
However, these methods are not calibration-free, requiring texture- or organic carbon-specific retraining to maintain accuracy.  
This limitation arises from their reliance on a sequential modeling approach, which assumes one-directional influence among components~(\eg measuring moisture first, then nutrients).
Such simplification overlooks the bidirectional relationships between soil components, known as cross-component interference~(introduced in §\ref{sec_introduction}), which leads to model drift when soil properties vary.

\textbf{Preliminary Experiments.}
We evaluate Scarf~\cite{ding2024cost} for organic carbon~$C$ estimation and SoilCares~\cite{wang2024soilcares} for $NPK$ measurement. 
Detailed experimental settings are provided in Section~\ref{sec_exp_setting}.

\begin{figure}[t]
\subfigure[Scarf~\cite{ding2024cost}: Soil texture variation introduces carbon estimation errors.]{
        \includegraphics[width=.22\textwidth]{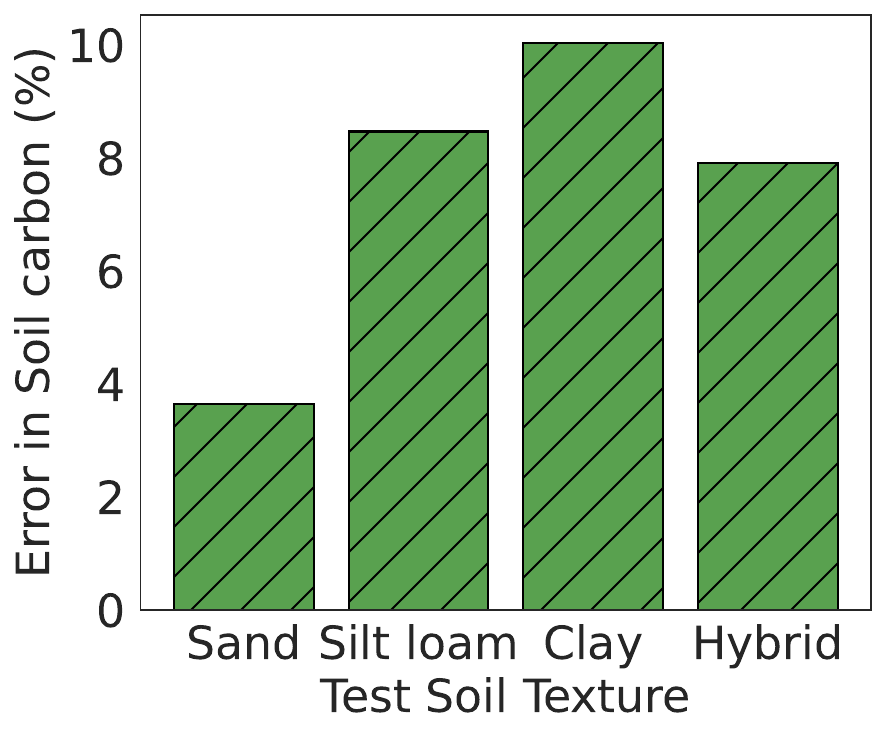}
        \label{fig_motivation_lora_dc}
    }
    \hspace{0.02in}
\subfigure[SoilCares~\cite{wang2024soilcares}: Carbon variation causes NPK estimation errors.]{
\includegraphics[width=.22\textwidth]{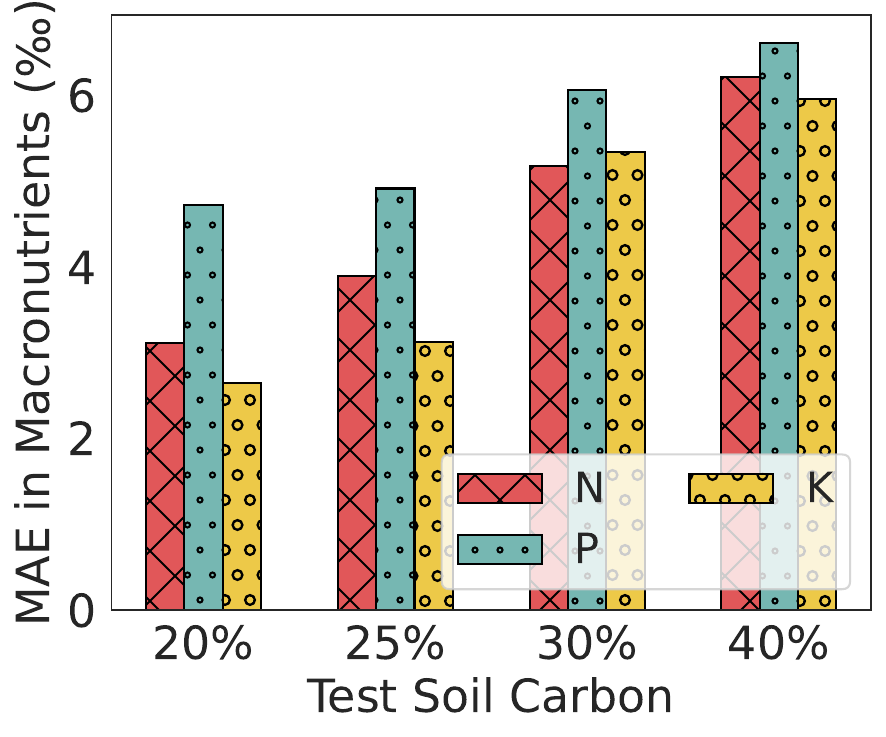}
        \label{fig_motivation_lora_inf}
    }
\caption{Cross-component interference observed in two state-of-the-art soil sensing systems~\cite{ding2024cost, wang2024soilcares}.}
  \Description[]{}
\end{figure}

For Scarf system~\cite{ding2024cost}, we replace the channel state information~(CSI)-derived dielectric permittivity with the permittivity calculated from our LoRa-based sensing. The model is trained using sand as the base soil texture and tested on soils with different textures~(\ie silt loam, clay, and a hybrid mix of 50\% clay, 25\% silt loam, and 25\% sand).
As shown in Figure~\ref{fig_motivation_lora_dc}, the carbon estimation error increases, rising from 3.65\% for the base texture to 8.47\%, 10.03\%, and 7.92\% for silt loam, clay, and hybrid textures, respectively.

Similarly, for SoilCares~\cite{wang2024soilcares}, we train the model on samples with 20\% soil carbon and test it on samples with carbon levels ranging from 25\% to 40\%.  
Figure~\ref{fig_motivation_lora_inf} demonstrates that macronutrient measurement errors are heavily influenced by carbon levels.  
Both experiments verify that existing multi-component soil sensing systems fail to account for cross-component interactions.

\subsection{Lack of E2U in LoRa Moisture Sensing}\label{sec_motivation_e2u}

\begin{figure}[tp]
\centering
    \subfigure[Ideal device placement.]{
    \label{fig_2antenna_correct}        \includegraphics[width=.41\linewidth]{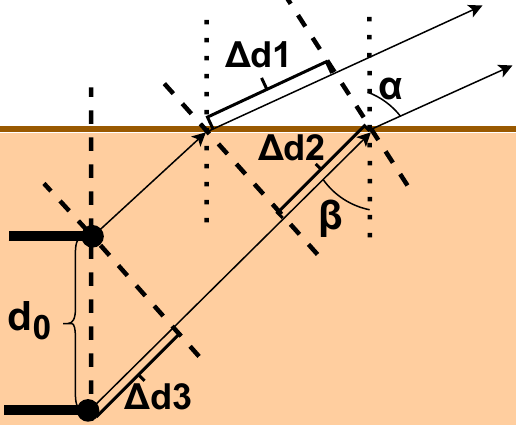}}
        \hfill
    \subfigure[Actual device placement.]{
    \label{fig_2antenna_wrong}
\includegraphics[width=.45\linewidth]{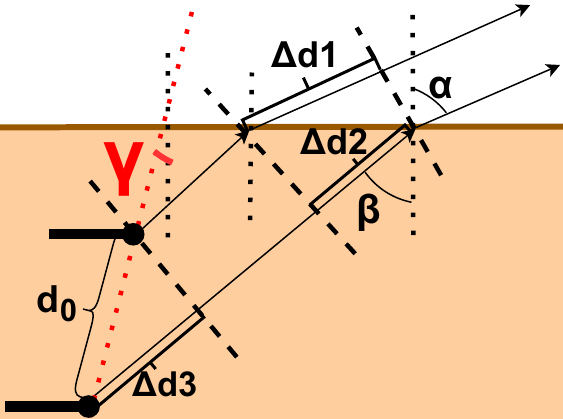}}
    \caption{Antenna placement, where the angle~$\beta$ typically varies within $\pm10^\circ$, based on Snell's law and the permittivity difference between air and soil. 
In practice,~$\beta$ is typically approximated as zero~\cite{chang2022sensor, wang2024soilcares}. 
(a) Ideal placement with antennas perpendicular to the soil surface. 
(b) Actual placement with a misalignment characterized by a rotation angle~$\gamma$.}
\Description[]{}
    
\end{figure}

Prior works~\cite{chang2022sensor, wang2024soilcares} demonstrate that LoRa signals can accurately measure soil moisture by computing phase shifts in dielectric permittivity~\(\epsilon\) via a dual-antenna switch design.  
As depicted in Figure~\ref{fig_2antenna_correct}, it consists of a buried LoRa node with two antennas and an above-ground gateway.  
This sensing method exploits the relationship between~\(\epsilon\) and volumetric water content~\(M\)~\cite{chang2022sensor, wang2024soilcares}, where~\(\epsilon\) correlates with the phase shift \(\Delta \phi\) between the two antennas:
\begin{equation}\label{eqn_moisture}
M= 0.1138\sqrt{\epsilon} - 0.1758 \;,
\end{equation}
\begin{equation}\label{eqn_phase_shift}
\Delta \phi = \frac{2 \pi f_c \Delta d_3}{c_0} \cdot \sqrt{\epsilon} = \frac{2 \pi f_c \sqrt{\epsilon}}{c_0} \cdot  \left(d_0 \cos \beta \right) \;,
\end{equation}
where the term~$\Delta d_3 = d_0 \cos \beta$~is the projection of the antenna distance~$d_0$ onto the LoRa transmission direction, defined by the angle~$\beta$, as illustrated in Figure~\ref{fig_2antenna_correct}.
The term $f_c$ denotes the LoRa frequency, and $c_0$ is the speed of light in a vacuum.

\textbf{Sensitivity to Antenna Placement.}  
According to Equation~\eqref{eqn_phase_shift}, accurate phase shift measurements depend on precise $\Delta d_3$ estimation.  
To compute~$\Delta d_3$, determining the angle~$\beta$ is necessary.  
Existing methods either manually measure this angle during initial deployment~\cite{ding2019towards} or assume it to be zero,~\ie $\Delta d_3 \approx d_0$~(assuming the line connecting the two antennas is perpendicular to the soil surface, as in Figure~\ref{fig_2antenna_correct}~\cite{chang2022sensor}.
However, both methods are unreliable.  
First, variations in soil permittivity due to soil dynamics continuously alter the angle.  
Second, soil movement may shift antenna placements, further distorting $\Delta d_3$ estimation.

To analyze this limitation, we examine the soil dielectric permittivity $\epsilon$ and soil moisture measurement errors caused by the antenna rotation angle $\gamma$ (the angle between expected and actual directions).  
As shown in Figure~\ref{fig_2antenna_wrong}, while the transmission times between $\Delta d_1$ and $\Delta d_2$ remain constant, $\Delta d_3$ shifts from $d_0 \cos\beta$ to $d_0 \cos\left(\beta - \gamma\right)$.  
This deviation results in a moisture error ratio:  
\begin{equation}
    \frac{\epsilon_\text{measure}}{\epsilon_\text{true}} = \cos^2\left(\beta-\gamma\right) \approx \cos^2\left(\gamma\right) \;.
\end{equation}
Considering Snell’s law and typical soil permittivity~$\epsilon$~(\eg 3–40), references~\cite{chang2022sensor, wang2024soilcares} approximate~$\beta$ as zero, thereby simplifying the error ratio to~$\cos^2\left(\gamma\right)$.
Consequently, as \(\gamma\) increases, the measurement error rises, peaking when the error ratio reaches zero.

\begin{figure}
\centering
{\includegraphics[width=0.46\textwidth]{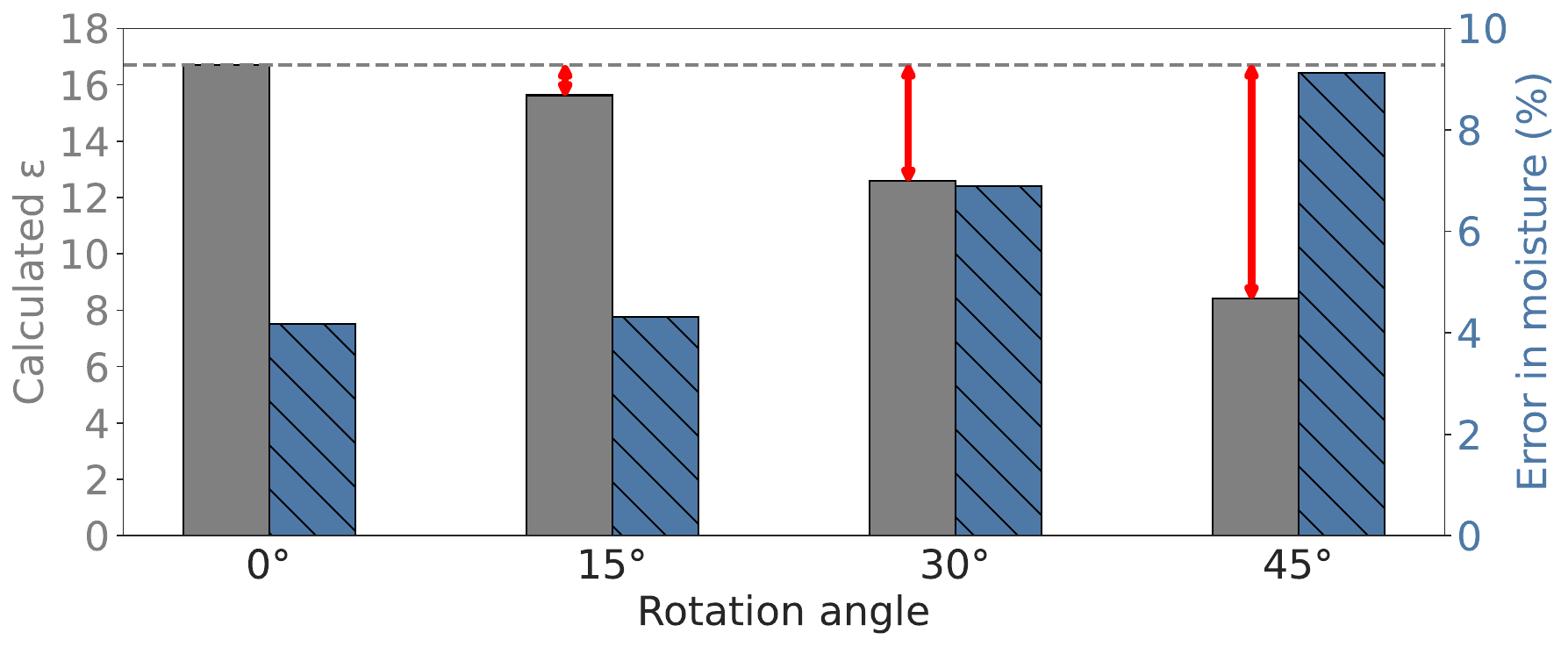}}
\caption{Impact of rotation angle $\gamma$ on soil moisture sensing accuracy using the method in~\cite{chang2022sensor}.}
\Description[]{}
    \label{fig_motivation_lora}
\end{figure}

\textbf{Preliminary Experiment.}
We conduct experiments to empirically evaluate the impact of the rotation angle~$\gamma$ on soil permittivity measurement~$\epsilon$.
Figure~\ref{fig_motivation_lora} illustrates how varying the rotation angle to 15, 30, and 45 degrees reduces the accuracy of the calculated $\epsilon$ using the method in~\cite{chang2022sensor}.  
This reduction leads to moisture estimation errors increasing by 3.6\%, 65.4\%, and 118.7\%, respectively.  
This highlights the sensitivity of existing systems to antenna alignment.

\section{{Related Work}}\label{sec_relatedWork}

\textbf{Soil Moisture Sensing.}
Various soil moisture sensing systems leverage different communication signals~\cite{ding2019towards, wang2020soil, chang2022sensor, khan2022estimating, feng2022lte, jiao2023soiltag, ding2023soil, ren2024demeter, chen2024metasoil}. 
SoilId~\cite{ding2023soil} sets up a metal plate in the soil and infers moisture with reflected UWB signal.
CoMEt~\cite{khan2022estimating} leverages soil layer reflection to infer permittivity above the soil surface. 
The high-bandwidth requirement and short sensing range restrict their adoption.
MetaSoil employs passive mmWave metamaterials for calibration-free, multi-layered sensing~\cite{chen2024metasoil}.
However, they neglect the influence of other components, such as organic carbon and soil texture, on moisture measurements~\cite{topp1980electromagnetic}.
\ourSystem addresses this by modeling cross-component interactions via 3CL framework, ensuring accurate estimation of multiple soil components. 
Additionally, unlike prior work that requires precise placement, \ourSystem's tetrahedral antenna design supports arbitrary placement, enhancing practicality.

\begin{table}[t]
\caption{Comparison of existing soil sensing systems (CF: Calibration-free, E2U: Easy-to-use).}
\centering
\begin{tabular}{L{0.8in}C{1.0in}C{0.5in}C{0.5in}}
\arrayrulecolor{black}\toprule
\textbf{System}   & \textbf{Comprehensiveness} & \textbf{CF} & \textbf{E2U} \\
\arrayrulecolor{black}\midrule
Strobe~\cite{ding2019towards}  & $M$ & $\times$ & $\times$ \\
\hline
GreenTag~\cite{wang2020soil}  & $M$ & $\times$ & $\times$ \\
\hline
Chang \etl~\cite{chang2022sensor}   & $M$ & $\times$ & $\times$ \\
\hline
CoMEt~\cite{khan2022estimating}  & $M$ & $\times$ & $\checkmark$ \\
\hline
LTE-Soil~\cite{feng2022lte}  & $M$ & $\times$ & $\times$ \\
\hline
SoilId~\cite{ding2023soil}    & $M$ & $\times$ & $\times$ \\
\hline
Demeter~\cite{ren2024demeter}    & $M$ & $\times$ &  $\times$ \\
\hline
MetaSoil~\cite{chen2024metasoil}  & $M$ & $\checkmark$ & $\checkmark$\\
\hline
SoilCares~\cite{wang2024soilcares}  & $M,N,P,K$ & $\times$ & $\times$ \\
\hline
Scarf~\cite{ding2024cost} & $M,C$ & $\times$ & $\times$ \\
\hline
\rowcolor{gray!20} \ourSystem   & $M,N,P,K,C,Al$ & $\checkmark$ & $\checkmark$ \\
\arrayrulecolor{black}\bottomrule
\label{table_comparison}
\end{tabular}
\end{table}

\begin{figure*}[t]
 \subfigure[Dielectric permittivity.]{
\includegraphics[width=.242\textwidth]{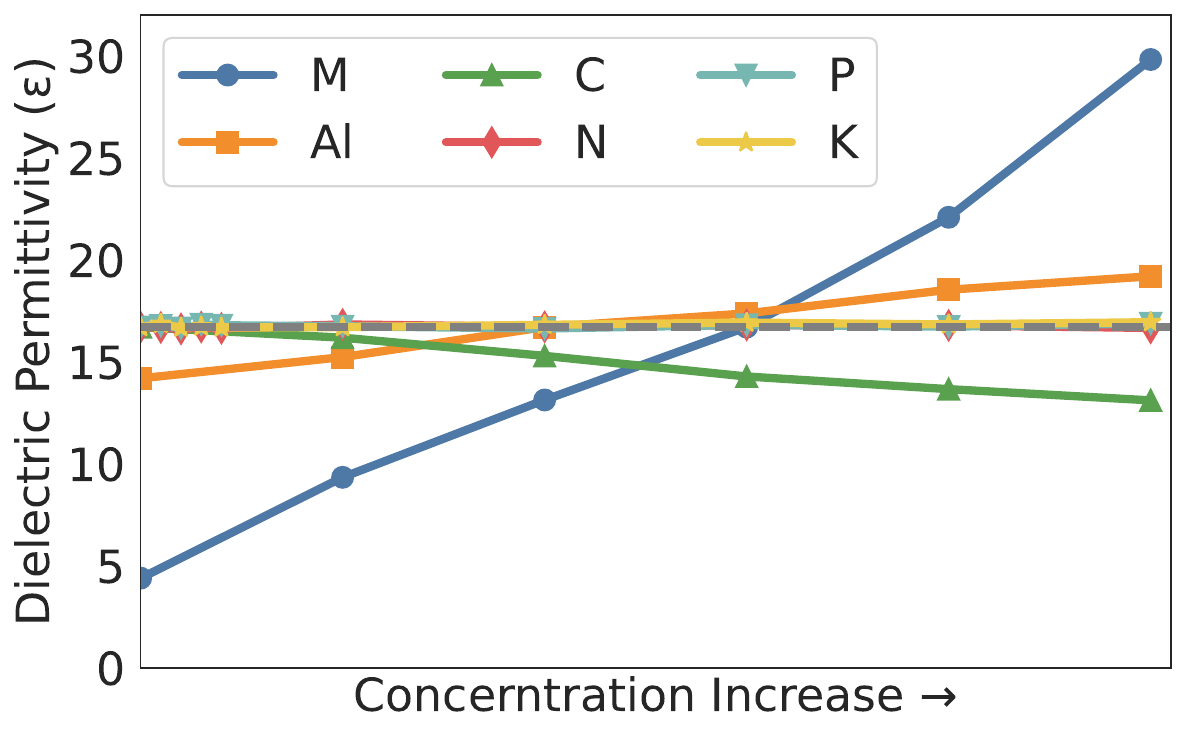}\label{fig_absorption_a}}
 \subfigure[Absorption for 1200 nm.]{
\includegraphics[width=.242\textwidth]{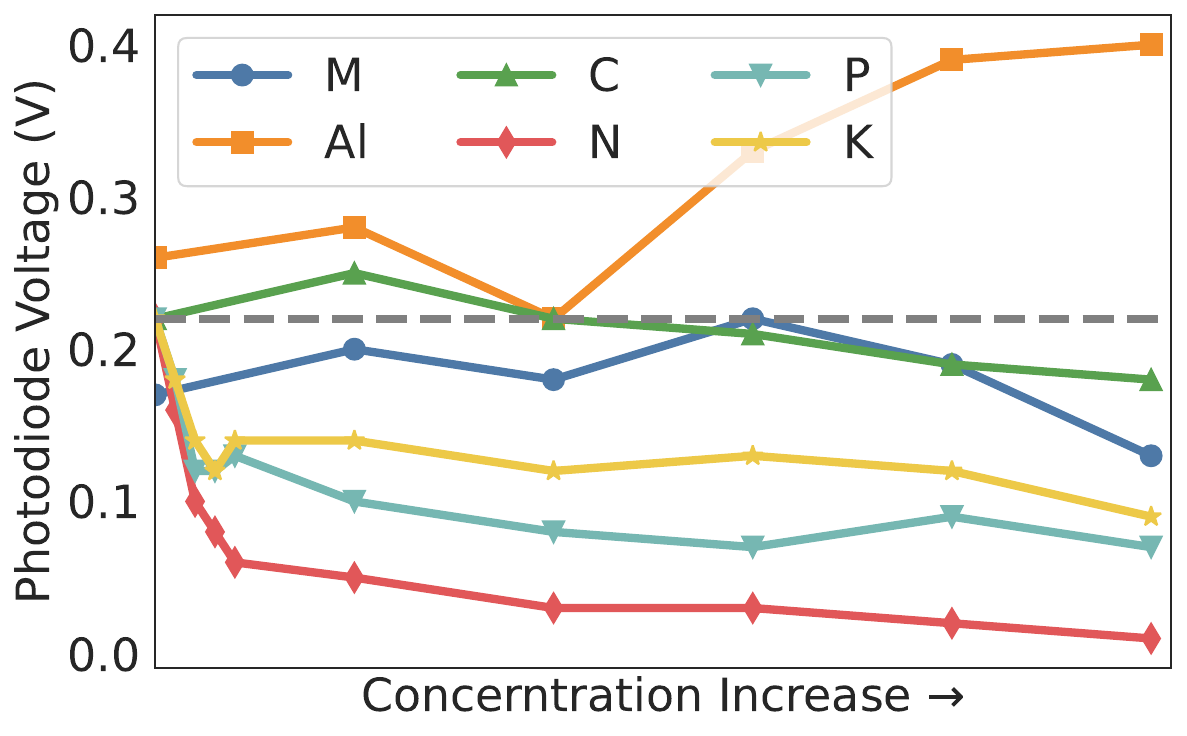}\label{fig_absorption_b}}
 \subfigure[Absorption for 1450 nm.]{
\includegraphics[width=.242\textwidth]{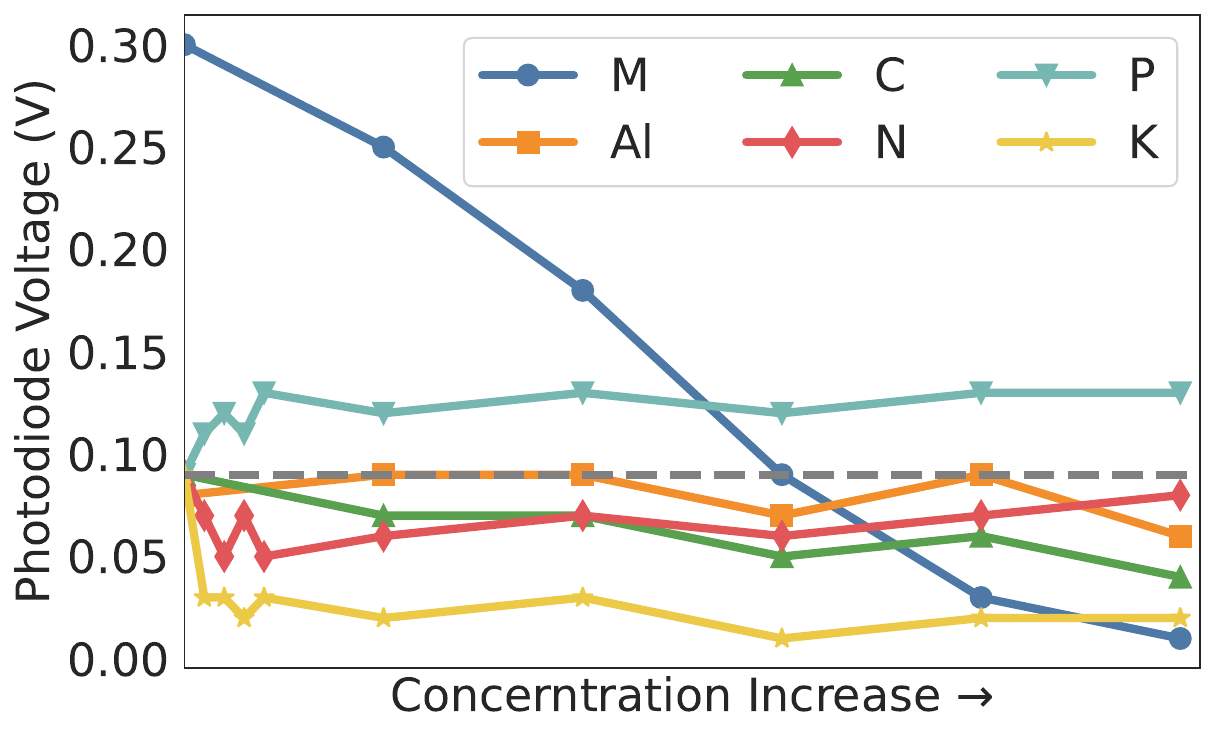}\label{fig_absorption_c}}
 \subfigure[Absorption for 1650 nm.]{
\includegraphics[width=.242\textwidth]{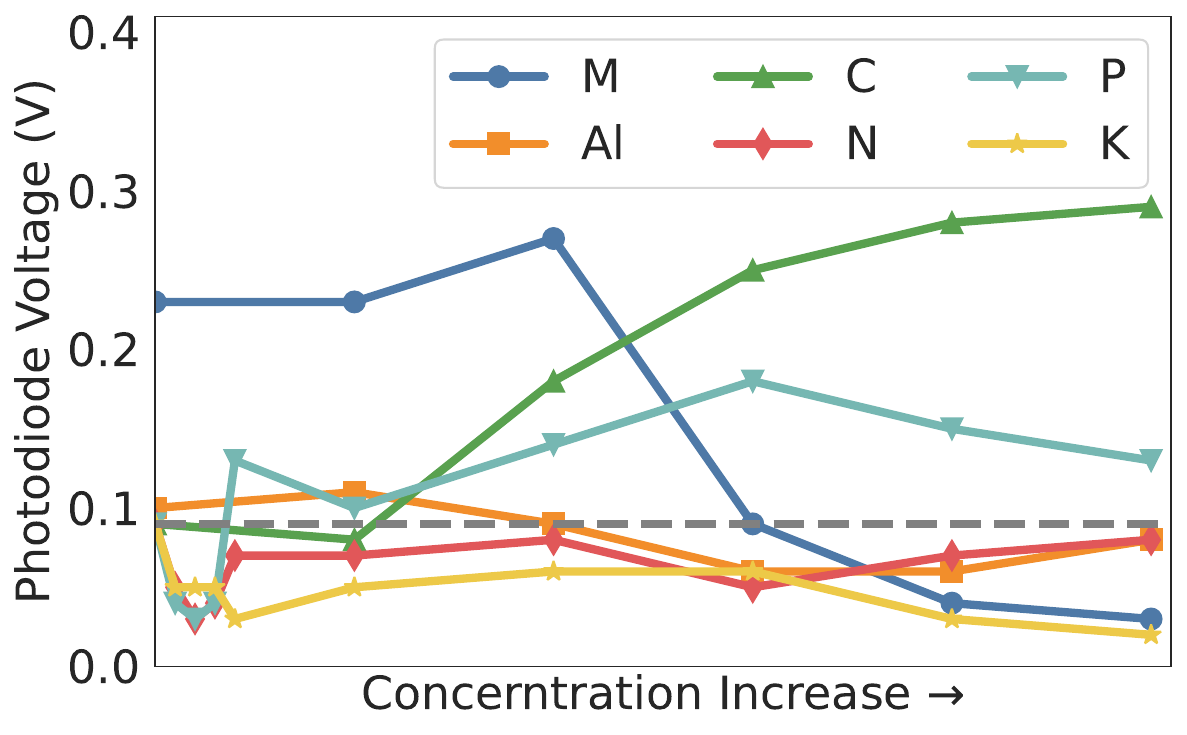}\label{fig_absorption_d}}
\caption{Permittivity and absorption features at three selected wavelengths as each soil component value varies.}
 \Description[]{On-site scenarios.}
    \label{fig_absorption}
\end{figure*}

\textbf{Soil Macronutrient and Carbon Sensing.}  
Reflectance spectroscopy is adopted for~$NPK$ analysis~\cite{bansod2014near, dhawale2013predicting, demiss2020comparison, hutengs2019situ}, but requires costly spectrometers and laborious processing. 
SoilCares~\cite{wang2024soilcares} combines RF-VNIR sensing via sequential modeling, estimating moisture first and then~$NPK$, assuming one-directional interference.

Compared with SoilCares~\cite{wang2024soilcares}, \ourSystem expands sensing from four to six components, eliminating texture- and carbon-specific recalibration. 
It adopts a calibration-free~3CL framework that decouples cross-component interference~\cite{ben1995near, rawls2003effect} through orthogonality and separation losses, removing the need for per-field fine-tuning. 
A tetrahedral LoRa antenna array enables placement-agnostic operation by mitigating CFO and baseband errors. 
Under identical data, \ourSystem achieves lower estimation error on~$\{M,N,P,K\}$ and uniquely supports accurate inference of~$C$ and~$Al$. 
In both lab and in-field experiments, it maintains stable performance across soil types, orientations, and environmental conditions, demonstrating strong robustness and generalizability.

Similarly, traditional carbon sensing methods offer high accuracy but are time-consuming and expensive~\cite{nelson1996total, chatterjee2009evaluation}. 
Recent systems like Scarf~\cite{ding2024cost} leverage Wi-Fi CSI and images for moisture and carbon measurement but remain sensitive to environmental factors and are not calibration-free. 
\ourSystem integrates RF and VNIR sensing while modeling cross-component dependencies, enabling accurate, scalable carbon analysis without extensive calibration.

As summarized in Table~\ref{table_comparison}, \ourSystem outperforms prior systems by jointly sensing all major soil components, enabling calibration-free operation, arbitrary placement, and cross-component modeling for robust, comprehensive soil component analysis.

\textbf{Contrastive Learning for Wireless Sensing.}  
Contrastive learning has been applied to wireless sensing to address challenges such as limited data and domain shifts. 
For instance, Cosmo~\cite{ouyang2022cosmo} bridges~IMU and depth camera features with diverse encoders for low-data scenarios, while ContrastSense~\cite{dai2024contrastsense} employs domain and time labels to enhance generalization. 
However, these methods focus on classification tasks, which do not suit the regression nature of soil sensing. 
RnC~\cite{zha2023rank} extends contrastive learning to regression by ranking samples based on numerical labels. 
However, it focuses only on single-output regression and does not address multi-output inference as required in our soil sensing task.
\ourSystem introduces a~3CL framework that extends classical contrastive learning from classification to multi-output regression across correlated soil components~$\{M,N,P,K,C,Al\}$. 
Unlike binary similarity objectives, 3CL leverages continuous target distances to preserve fine-grained relationships among components. 
It further adds a \emph{Separation Loss} to reduce cross-component interference and an \emph{Orthogonality Regularizer} to decorrelate feature representations, enabling calibration-free and label-efficient learning across diverse soil conditions.

\section{{Design of \ourSystem}}
\label{sec_design}

\ourSystem incorporates two key modules: Contrastive Cross-Component Learning~(3CL) for calibration-free soil sensing, and a Tetrahedron Antenna Array for robust device placement.

\subsection{Overview}

\begin{figure}
\centering
{\includegraphics[width=0.49\textwidth]{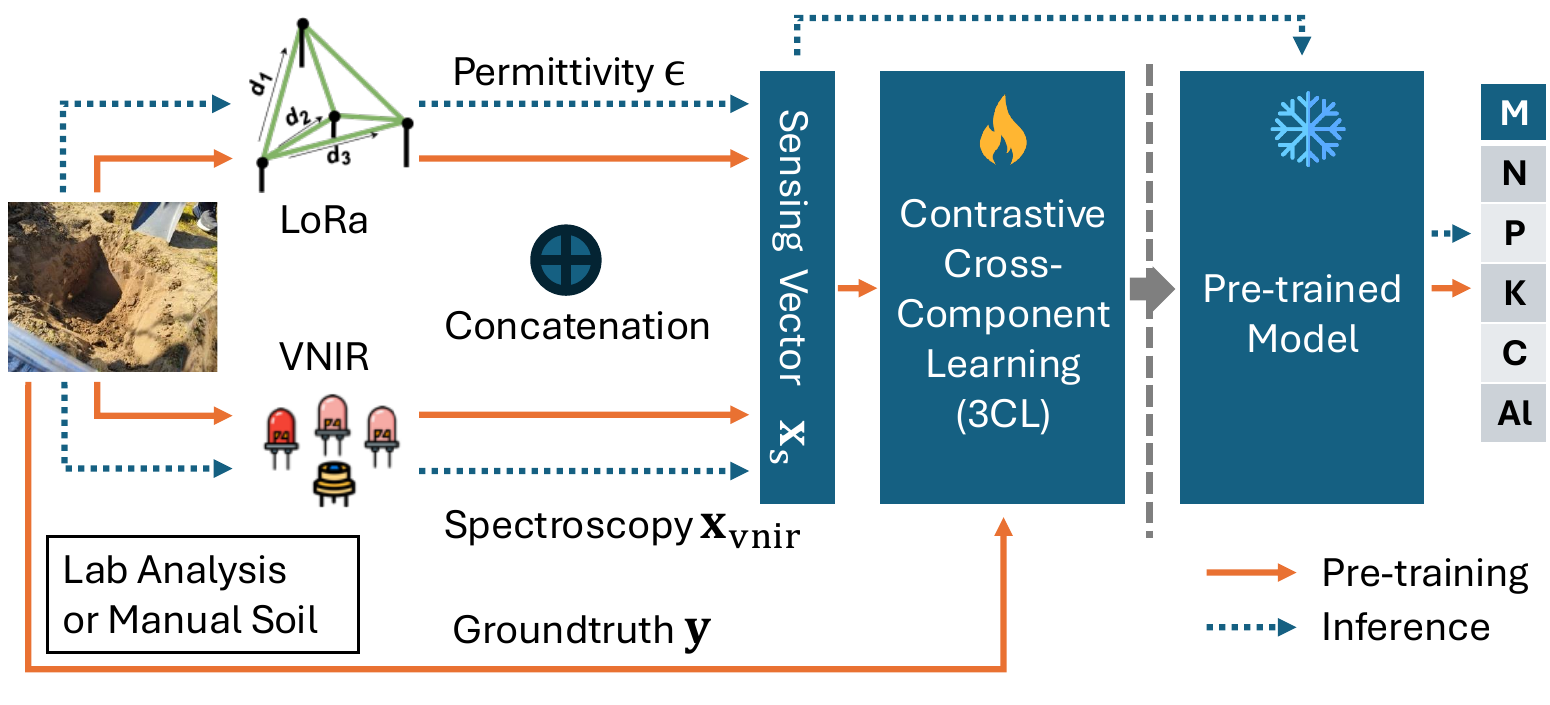}}
\caption{Workflow of \ourSystem. 
Spectroscopy captures the attenuation of multiple VNIR wavelengths reflected from a soil sample. 
\ourSystem focuses on six components: moisture~($M$), macronutrients~($NPK$), carbon~($C$), and aluminum~($Al$).}
\Description[]{}
\label{fig_work_flow}
\end{figure}

As illustrated in Figure~\ref{fig_work_flow}, \ourSystem fuses LoRa and VNIR channels to estimate six soil components~\(\{M, N, P, K, C, Al\}\).

\textbullet~\textbf{Tetrahedron-Based LoRa Channel for Permittivity~$\epsilon$.}
LoRa packets are transmitted from our tetrahedral four-antenna setup to measure soil permittivity $\epsilon$~(detailed in~§\ref{sec_tetrahedron}).  
This enables the system to derive information about $M$, $C$, and $Al$, since permittivity~$\epsilon$ is influenced by these factors together~\cite{gharechelou2020mineral, abdulraheem2024recent}.

\textbullet~\textbf{VNIR Channel for Spectroscopy~$\mathbf{x}_\text{vnir}$.}
We strategically select seven VNIR wavelengths that are sensitive to specific soil components.
After each light wave is sent out and then reflected from the soil sample, we obtain the corresponding attenuation value.
Together, the six attenuation values form the spectroscopy vector~$\mathbf{x}_\text{vnir}$, which characterizes the components~\(\{M, N, P, K, C, Al\}\).

The soil permittivity~$\epsilon$ and spectroscopy features~$\mathbf{x}_\text{vnir}$ are then concatenated to form the sensing vector~\(\mathbf{x}_{s} = \left[\epsilon;\ \mathbf{x}_\text{vnir}\right]\).

\textit{i)~For the pre-training stage,} the ground truth of the six soil components,~$\mathbf{y} = \left[ y_M, y_N, y_P, y_K, y_C, y_{Al} \right]^\top$, is required.
Obtaining these values typically involves costly in-laboratory soil analysis~\cite{van2000soil} or manual sample preparation.  
In this work, we manually formulate a small number of soil samples for training~(details in §\ref{sec_otho}).
Each soil sample's sensing vector~$\mathbf{x}_{s}$ and its corresponding ground truth vector~$\mathbf{y}$ are then input into the~3CL module~(§\ref{sec_contrastive_learning}) for accurate multi-component estimation.

\textit{ii)~For the inference stage,} given a new soil sample, either manually formulated or collected from the field, the measured sensing vector~\(\mathbf{x}_\text{s}^\text{new}\) is fed into the pre-trained model to predict the six soil component values:~$\hat{\mathbf{y}} = \left[ \hat{y}_M, \hat{y}_N, \hat{y}_P, \hat{y}_K, \hat{y}_C, \hat{y}_{Al} \right]$.

\subsection{Permittivity \& Spectroscopy}

Before detailing the design of \ourSystem, we first explain why the combined measurements of permittivity~\(\epsilon\) and spectroscopy~\(\mathbf{x}_\text{vnir}\) are sufficient to quantify six soil components~\(\{M, N, P, K, C, Al\}\).

\subsubsection{Unique Responses and Complementary Sensing Channels.}
Each sensing channel in \ourSystem, whether permittivity or VNIR spectroscopy, responds uniquely to specific soil components, albeit with some overlap. 
Soil dielectric permittivity~\(\epsilon\) is highly sensitive to~$M$, moderately influenced by~$C$ and~$Al$, and minimally affected by macronutrients~$NPK$, as illustrated in Figure~\ref{fig_absorption_a}.
Specifically, Figure~\ref{fig_absorption_a} shows that as the concentration of~$M$ increases, the permittivity rises sharply from approximately 5 to 30, while~$C$ and~$Al$ cause smaller increases (from 5 to around 10–15). 
In contrast,~$NPK$ have negligible effects, with values remaining near 15. 
This distinct sensitivity profile allows permittivity to primarily capture moisture, while also providing secondary information about~$C$ and~$Al$.

VNIR spectroscopy complements permittivity by targeting~$NPK$, while also enhancing the detection of all components. 
We select seven VNIR wavelengths to capture the absorption characteristics of each component: 1450~nm for~$M$, 1650~nm for~$C$, 1200~nm for~$N$, 620~nm for~$P$, and 460~nm for~$K$, along with two additional wavelengths (1300~nm and 1550~nm) to capture combined effects. 
The absorption spectrum of~$Al$ spans broadly from 250~nm to 2500~nm, allowing it to be indirectly captured through these wavelengths.

Figures~\ref{fig_absorption_b}–\ref{fig_absorption_d} highlight the distinct absorption patterns at three wavelengths. 
At 1200~nm~(Figure~\ref{fig_absorption_b}), increasing~$N$ reduces the photodiode voltage from 0.22~V to 0.01~V, while~$Al$ increases it from 0.25~V to 0.39~V, demonstrating opposing trends that aid in distinguishing these components. 
At 1450~nm~(Figure~\ref{fig_absorption_c}),~$M$ causes a sharp voltage drop from 0.30~V to 0.02~V, while~$C$ and~$Al$ show minimal changes (remaining around 0.20–0.25~V). 
At 1650~nm~(Figure~\ref{fig_absorption_d}),~$C$ reduces the voltage from 0.30~V to 0.10~V, with~$M$ and~$Al$ having smaller effects (around 0.20–0.25~V). 
These distinct spectral responses, combined with permittivity, provide a strong foundation for accurate component quantification.

\textbf{Component-Specific VNIR Absorption and Detector Selection.}
To isolate single-component effects, we vary one soil component at a time and measure photodiode voltages at 1200, 1450, and 1650\,nm using detectors with flat responsivity across 1.0--1.7\,$\mu$m. 
The measurements reveal clear signatures: nitrogen reduces voltage at 1200\,nm, moisture causes strong absorption at 1450\,nm, and organic carbon increases absorption at 1650\,nm, while aluminosilicate texture shows minimal influence. 
These patterns justify the wavelength–detector pairing: 1450\,nm for moisture, 1650\,nm for organic carbon, and 1200\,nm for nitrogen. 
Si photodiodes are used for visible bands (460 and 620\,nm) and InGaAs photodiodes for 1200--1650\,nm to ensure uniform NIR responsivity.

\begin{figure}
\centering
{\includegraphics[width=0.46\textwidth]{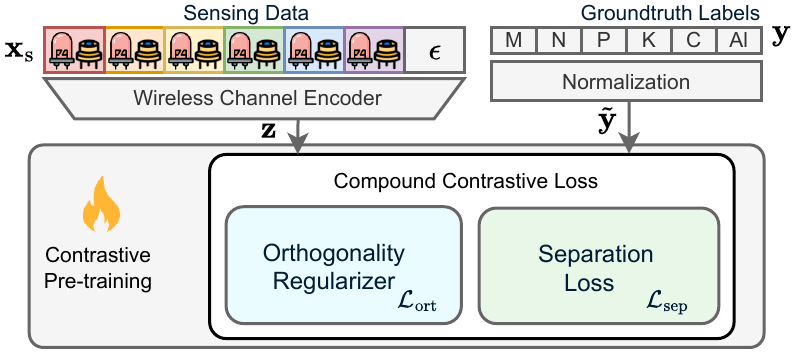}}
\vspace{1pt}
\caption{Illustration of \ourSystem pre-training workflow.}
\Description[]{}
\label{fig_model_sys}
\end{figure}

\subsubsection{Small Voltage Differences.}
A potential concern arises from the fact that some VNIR measurements (Figures~\ref{fig_absorption_b}–\ref{fig_absorption_d}) show less than 0.1~V variation across component concentrations, raising doubts about separability when signals are subtle or correlated.

However, \ourSystem is designed such that each of the seven selected VNIR wavelengths primarily responds to a specific component: 1450~nm for~$M$, 1650~nm for~$C$, 1200~nm for~$N$, 620~nm for~$P$, 460~nm for~$K$, with 1300~nm and 1550~nm capturing broader effects, including~$Al$.
This component-specific mapping ensures that strong sensitivity at a single wavelength is sufficient. 
Minor shifts caused by other components at that wavelength are expected and do not impair disentanglement.
For example, at 1450~nm,~$M$ induces a strong voltage drop~(from 0.30~V to 0.05~V), while~$C$ and~$Al$ remain nearly flat. 
Note that Figures~\ref{fig_absorption} only illustrate three of the seven wavelengths.
Moreover, permittivity~(\(\epsilon\)) and VNIR signals~(\(\mathbf{x}_\text{vnir}\)) reflect distinct fundamental physical properties:~\(\epsilon\) captures the bulk dielectric response, primarily from~$M$, while VNIR captures molecular absorption. 
Their complementary nature enhances identifiability.

\textbf{Summary.}
The combined measurements of permittivity~\(\epsilon\) and VNIR spectroscopy~\(\mathbf{x}_\text{vnir}\) enable quantification of six soil components. 
Permittivity mainly captures~$M$, with secondary sensitivity to~$C$ and~$Al$, while VNIR targets~$NPK$ and supports all components via selected wavelengths. 
Each of the seven wavelengths, three shown, aligns with a primary component and leverages the complementary nature of dielectric and molecular absorption.

\subsection{Contrastive Cross-Component Learning}\label{sec_contrastive_learning}

Figure~\ref{fig_model_sys} depicts the proposed 3CL module.  
It begins with a wireless channel encoder that maps each soil sample’s sensing vector~\(\mathbf{x}_\text{s}\) into a $D$-dimensional embedding~(with $D = 512$) in the latent space.
This upscaling preserves intricate patterns in the sensing data, enabling subsequent stages to disentangle soil component representations.

\begin{figure}
\centering
\subfigure[Before training.]{
\label{fig_vis_a}       \includegraphics[width=.29\linewidth]{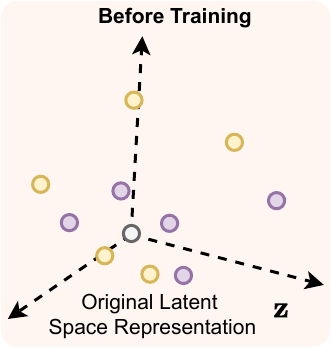}}
\subfigure[Only orthogonality.]{
\label{fig_vis_b}
\includegraphics[width=.32\linewidth]{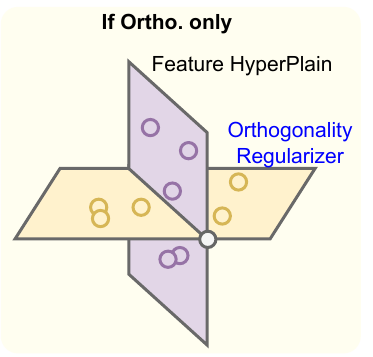}}
\subfigure[Compound loss.]{
\label{fig_vis_c}
\includegraphics[width=.33\linewidth]{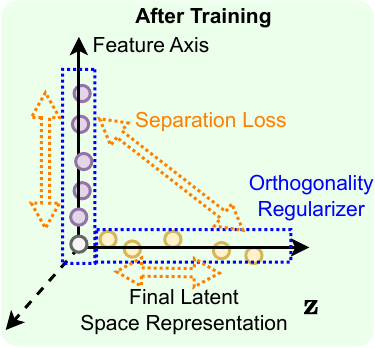}}
\caption{Demonstration of the latent space before and after training using Compound Contrastive Loss.}
\Description[]{}
\label{fig_model_vis}
\end{figure}

After mapping each soil sample’s sensing vector~\(\mathbf{x}_{s} = \left[\epsilon;\ \mathbf{x}_\text{vnir}\right]\) from the training dataset into the latent space, the resulting representations appear initially scattered, as illustrated in Figure~\ref{fig_vis_a}.
To effectively structure this latent space, we employ the Compound Contrastive Loss in 3CL, which integrates the Orthogonality Regularizer and the Separation Loss.  
This loss function optimizes latent space representations by achieving two key objectives:

\textbf{i)~From Soil Component Perspective.}
The Orthogonality Regularizer encourages the effects of different soil components in the latent space to remain independent, facilitating the disentanglement of each component's impact on the sensing signals.
As shown in Figure~\ref{fig_vis_b}, where distinct color planes (\eg purple and yellow) represent the impact of different soil components, the Orthogonality Regularizer organizes the latent space into corresponding planes for each component.
However, when applied alone, it may not sufficiently separate samples with different ground truth component values~$\mathbf{y}$, resulting in overlap within each latent space plane.

\textbf{ii)~From Soil Sample Perspective.}
The Separation Loss differentiates soil samples based on their ground truth component values~$\mathbf{y}$, which are normalized to~$\tilde{\mathbf{y}}$.
It encourages distances between sample representations in the latent space to align with differences in their ground truth component values.  
When combined with the Orthogonality Regularizer, it helps structure the latent space effectively, as shown in~Figure~\ref{fig_vis_c}. 
Here, samples are clearly separated along different axes~(\eg purple and yellow), while preserving the independence of soil component effects.

\subsubsection{Orthogonality Regularizer}
\label{sec_otho}

To disentangle the effects of soil components~$\left\{M, N,P,K, C, Al\right\}$ in the latent space, each component’s influence should be represented independently, minimizing overlap with others.  
To achieve this, the Orthogonality Regularizer follows two steps:  
i)~identifying the effect (\ie direction) of each component in the latent space,  
ii)~orthogonalizing these directions.

\textbf{i)~Identifying Each Component Direction.}  
We design a training dataset with a controlled number of soil samples to determine each component's direction in the latent space.  
First, a reference soil sample with typical values~$\{M$=$30\%, Al$=$4\%, N$=$0, P$=$0, K$=$0, C$=$0\}$ is selected, with its embedding vector denoted as \(\mathbf{z}^0\).

\textsc{Training Data.}
Controlled soil samples are manually generated by varying one component at a time while keeping the other five fixed. 
Table~\ref{tab_training_data_small} presents examples of the reference sample and its variations with changing~$M$ or~$N$ values.
For instance, one sample is created by adjusting the reference sample’s~$M$ value from~30\% to~10\%, with all other components unchanged.
In total, 43 soil samples are generated.
Specifically, each component’s variation range is predefined:~$M$ ranges from~0\% to~50\% in~10\% steps;~$Al$ from~0\% to~10\% in~2\% steps;~$C$ from~10\% to~50\% in~10\% steps; 
and~$NPK$ from~0.02\% to~0.08\% in~0.02\% steps, then from~0.2\% to~1\% in~0.2\% steps.
Although the step sizes differ across components, they do not constrain~\ourSystem’s resolution, as our regression-based sensing approach can interpolate between sampled values.

All formulated samples are categorized into six component-specific groups~(\eg $M$-group, $N$-group, $Al$-group, \textit{etc.}), where each sample is assigned to the group corresponding to the component that differs from the reference soil sample.
For example, the $M$-group includes all controlled soil samples generated above where only the~$M$ value deviates from the reference soil sample.

Within a group, every soil sample is mapped to a~$D$-dimensional embedding vector~\(\mathbf{z}_g^i\) via the wireless channel encoder, where~\(g \in \{M, N, P, K, C, Al\}\) denotes the component-specific group, and~\(i\) represents the sample index within that group.
We then compute the average direction of each component’s effect in the latent space by comparing the embedding vectors of the group’s soil samples~\(\mathbf{z}_g^i\) to the reference sample’s embedding~\(\mathbf{z}^0\):
\begin{equation}
\mathbf{z}_g^{\text{avg}} = \frac{1}{n_g} \sum_{i=1}^{n_g} \left(\mathbf{z}_g^i - \mathbf{z}^0\right) \;,
\end{equation}
where \(n_g\) is the number of samples in the group~$g$~(\eg 5 samples in $M$-group).
The embedding vector~\(\mathbf{z}^{\text{avg}}_g\) captures how the component~\(g\) influences the embedding, as only the values of~\(g\) vary within each group while all other components remain fixed.
Thus, the change~\(\left(\mathbf{z}_g^i - \mathbf{z}^0\right)\) isolates the effect of component~\(g\) on the latent representation, since only~\(g\)'s value varies within the group while all other component values remain constant.
By averaging these changes across all samples in the group, we obtain a stable and representative directional shift~\(\mathbf{z}_g^{\text{avg}}\) in the latent space, which captures the influence of the soil component~\(g \in \{M, N, P, K, C, Al\}\).

\textbf{ii)~Orthogonalizing Directions.}  
After computing each soil component's directional vector~\(\mathbf{z}^{\text{avg}}_g\) in the latent space, they may still overlap due to cross-component interference.
To address this, we enforce orthogonality among component directions, minimizing overlap similar to the x, y, and z axes in 3D space.
We define the Orthogonality Regularizer as follows:
\begin{equation}
\mathcal{L}_{\text{ort}} = \left\| \mathbf{Z}^T \mathbf{Z} - \mathbf{I} \right\|_F^2, \quad
\mathbf{Z} = \begin{bmatrix} \mathbf{z}^{\text{avg}}_{M}  & \dots & \mathbf{z}^{\text{avg}}_{Al} \end{bmatrix}^T \in \mathbb{R}^{6 \times D} \;,
\end{equation}
where~\(\mathbf{Z} \in \mathbb{R}^{6 \times D}\) is a matrix with each row representing the~\(\mathbf{z}^{\text{avg}}_{g}\) vector for each soil component, \(\mathbf{Z}^T \mathbf{Z} \in \mathbb{R}^{6 \times 6}\) quantifies the overlap between these directions, \(\mathbf{I} \in \mathbb{R}^{6 \times 6}\) is the identity matrix~(indicating perfect orthogonality), and \(\| \cdot \|_F\) measures the difference between matrices.  
By minimizing the loss term~\(\mathcal{L}_{\text{ort}}\), the directions corresponding to~\(\{M, N, P, K, C, Al\}\) become more independent, reducing the risk of their effects mixing in the latent space.

\begin{table}[t]
\caption{Examples of controlled soil samples for training,~\eg~varying~$M$ while keeping other components fixed.}
    \centering
    \renewcommand\arraystretch{1.0}
    \begin{tabular}{C{0.5in}C{0.29in}C{0.29in}C{0.29in}C{0.29in}C{0.29in}C{0.29in}C{0.29in}}
        \toprule
         & $M$ & $Al$ & $C$ & $N$ & $P$ & $K$ \\
        \midrule
        \rowcolor{gray!20} Reference & 30\% & 4\% & 0\% & 0\% & 0\% & 0\% \\
        $M1$ & \textbf{10\%} & 4\% & 0\% & 0\% & 0\% & 0\% \\
         $N1$ & 30\% & 4\% & 0\% & \textbf{0.02\%} & 0\% & 0\% \\
        \bottomrule
    \end{tabular}
    \label{tab_training_data_small}
\end{table}

\subsubsection{Separation Loss}
While the Orthogonality Regularizer encourages the effects of soil components to be independent in the latent space, we also control the relative distances between latent representations of different soil samples.
The Separation Loss achieves this goal by aligning the distances between latent embedding vectors with the differences in their ground truth component values for any two given soil samples.

Specifically, for any pair of soil samples~\(\mathbf{x}_\text{s}^i\) and~\(\mathbf{x}_\text{s}^j\) with their normalized labels \(\tilde{\mathbf{y}}^i\) and \(\tilde{\mathbf{y}}^j\), we define the loss as:
\begin{equation}
\mathcal{L}_{\text{sep}} = \sum_{i \neq j} \left( \|\mathbf{z}^i - \mathbf{z}^j\|_2 - \|\tilde{\mathbf{y}}^i - \tilde{\mathbf{y}}^j\|_2 \right)^2 \;,
\end{equation}
where~$\mathbf{z}^i$ and~$\mathbf{z}^j$ represent the embedding vectors of the soil samples~\(\mathbf{x}_\text{s}^i\) and~\(\mathbf{x}_\text{s}^j\), respectively, and~\(\|\cdot\|_2\) denotes the Euclidean~(L2) norm. 
Given~\(L\) training soil samples, a total of~\(\binom{L}{2}\) unique pairs are formed for this loss calculation.

This loss encourages the Euclidean distance between latent embeddings~\(\|\mathbf{z}^i - \mathbf{z}^j\|_2\) to match the distance between their normalized labels~\(\|\tilde{\mathbf{y}}^i - \tilde{\mathbf{y}}^j\|_2\).  
For example, if two samples have distinct moisture levels~(\eg \(M = 10\%\) vs. \(M = 40\%\)), their label distance will be large, and the loss pushes their embeddings farther apart in the latent space.  
By aligning these distances, the Separation Loss helps the encoder differentiate samples with varying component values.

\subsubsection{Compound Contrastive Loss.}
To structure a latent space where the effects of soil components are both distinct and independent, we weighted combine the Separation Loss and Orthogonality Regularizer into a weighted final contrastive loss:
\begin{equation}
\mathcal{L} = \lambda_{\text{sep}}\mathcal{L}_{\text{sep}} + \lambda_{\text{ort}}\mathcal{L}_{\text{ort}} \;,
\end{equation}
where \(\mathcal{L}_{\text{sep}}\) separates samples based on label differences, and~\(\mathcal{L}_{\text{ort}}\) aligns component effects along orthogonal directions.  
The hyperparameters \(\lambda_{\text{sep}}\) and \(\lambda_{\text{ort}}\) balance these objectives, with their values determined through extensive hyperparameter tuning. 
In our current implementation,~$\lambda_{\text{sep}}$ and~$\lambda_{\text{ort}}$ are set to 0.82 and 0.18, respectively.

Training uses gradient-based optimization with regular checkpoints to mitigate overfitting, while input feature normalization and early stopping based on validation loss trends help maintain stable learning. 
This weighted loss effectively structures the latent space. 
As visualized in Figure~\ref{fig_model_vis}, before pre-training, the embeddings~$\mathbf{z}_i$ are randomly distributed in the space. 
After training with the compound contrastive loss, their spatial distribution becomes structured, enabling disentanglement of different components and precise inference of their numerical values.

\subsubsection{Inference}
After pre-training, we retain four stored representations:  
pre-trained wireless channel encoder,  
average direction vectors~\(\mathbf{z}_g^{\text{avg}}\) for each soil component,  
the reference soil sample’s embedding~\(\mathbf{z}^0\),  
and the recorded normalization coefficients for each component’s ground truth.

Given a new soil sample's measured sensing vector~\(\mathbf{x}_\text{s}^\text{new}\), we first compute its~\(D\)-dimensional embedding vector~\(\mathbf{z}^\text{new}\) using the pre-trained wireless channel encoder.
Next, we subtract the reference embedding vector to obtain the deviation:~\(\mathbf{z}^\text{new} - \mathbf{z}^0\). 
This deviation is then projected onto the six average direction vectors~\(\mathbf{z}_g^{\text{avg}}\) by computing the dot product for each:
\begin{equation}
    \text{score}_g = (\mathbf{z}^\text{new} - \mathbf{z}^0) \cdot \mathbf{z}_{g}^{\text{avg}}, \quad g \in \{M, N, P, K, C, Al\} \;.
\end{equation}
This computation yields six scalar values~\(\text{score}_g\), each representing the contribution of a specific soil component.
Finally, we reverse the normalization by applying the recorded coefficients to each~\(\text{score}_g\), recovering the predicted values for the six soil components in their original scale~(\eg moisture~$M$ in percentage).

\subsection{E2U Tetrahedron Antenna Array}\label{sec_tetrahedron}

To measure phase shift~$\Delta \phi$ for soil dielectric permittivity~$\epsilon$, the LoRa dual-antenna array, as introduced in §\ref{sec_motivation_e2u}, requires the line connecting the two antenna front points to be perpendicular to the soil surface~\cite{chang2022sensor}.  
To relax the strict placement requirement of the LoRa dual-antenna array, we propose a tetrahedral antenna array with an antenna switching mechanism, termed Easy-to-Use (E2U).

\subsubsection{Tetrahedral Antenna Array}

\begin{figure}
\centering
    \subfigure[Tetrahedron Placement.]{
    \label{fig_tera_antenna}
\includegraphics[width=.35\linewidth]{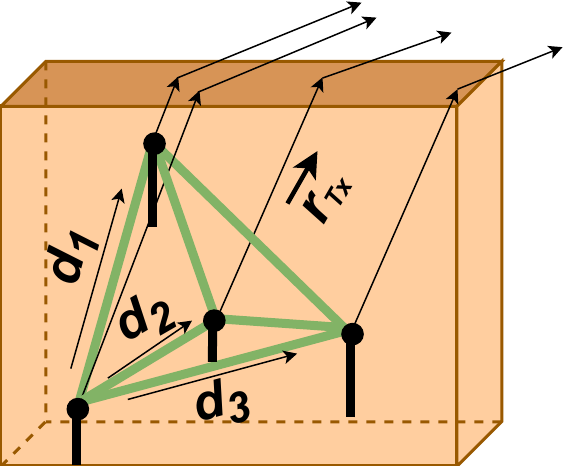}}
        \hfill
    \subfigure[SP4T RF switching in preambles.]{
    \label{fig_tera_switch}
\includegraphics[width=.615\linewidth]{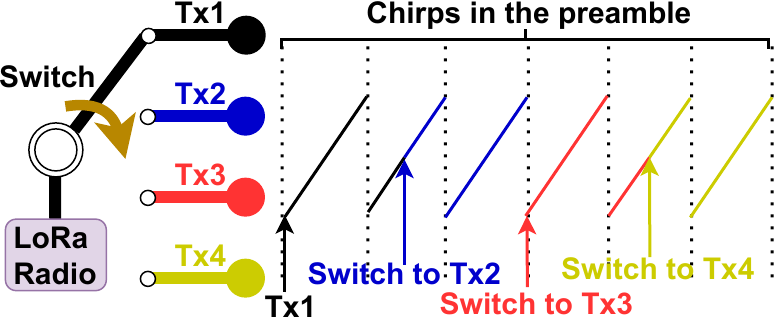}}
\caption{(a) Tetrahedron placement with four antennas in a 3D configuration; (b) Single pole four throw~(SPFT) RF switch controls antenna selection for preamble chirp transmission.}
\Description[]{}
\label{fig_trans}
\end{figure}

Figure~\ref{fig_tera_antenna} depicts the placement of our four LoRa antennas, positioned at the vertices of a tetrahedron with equal side lengths.
To understand why our tetrahedral array works, we reformulate Equation~\eqref{eqn_phase_shift} for phase shift calculation.

Specifically, the phase shift $\Delta \phi$ in a dual-antenna array is determined by the projection $\left(  d_0 \cos \beta = \vec{d} \cdot \vec{r_{\text{TX}}}\right)$, where $\vec{d}$ is the vector connecting the front points from the lower to the upper antenna, and $\vec{r_{\text{TX}}}$ is the LoRa signal transmission direction:
\begin{equation}
\Delta \phi = \frac{2 \pi f_c \sqrt{\epsilon}}{c_0} \left(\vec{d} \cdot \vec{r_{\text{TX}}}\right) \;,
\end{equation}

To calculate the projection~\(\left(\vec{d} \cdot \vec{r}_{\text{TX}}\right)\), it is essential to determine the angle between~\(\vec{d}\) and~\(\vec{r}_{\text{TX}}\), which is influenced by the rotation angle~\(\gamma\) in the dual-antenna array, as introduced in~§\ref{sec_motivation_e2u}.
Our tetrahedral antenna array leverages multiple LoRa transmission paths to eliminate the need to manually measure the rotation angle~\(\gamma\), instead of relying on a single path as in the dual-antenna array.

We randomly designate one vertex as the origin
within a Cartesian coordinate system, with the other three antennas located at:
\begin{equation}
\begin{aligned}
\vec{d_1} &= \left[ \frac{\sqrt{3}}{3} d_0, 0, -\frac{\sqrt{6}}{3} d_0 \right]  \;,\\
\vec{d_2} &= \left[ -\frac{\sqrt{3}}{6} d_0, -\frac{1}{2} d_0, -\frac{\sqrt{6}}{3} d_0 \right]  \;,\\
\vec{d_3} &= \left[ -\frac{\sqrt{3}}{6} d_0, \frac{1}{2} d_0, -\frac{\sqrt{6}}{3} d_0 \right] \;,
\end{aligned}
\end{equation}
where the term~$d_0$ is the distance between two antennas.  
Expanding Equation~\eqref{eqn_phase_shift}, we obtain three phase shifts~\(\{\Delta \phi_k\}_{k=1}^3\) from the origin antenna to the other three antennas, defined as follows:
\begin{equation}
\begin{aligned} \label{eqn_phase_shifts}
    \Delta \phi_k &= \frac{2 \pi f_c \sqrt{\epsilon}}{c_0} \left(\vec{d_k} \cdot \vec{r_{\text{TX}}}\right), \quad k \in \left\{1, 2, 3\right\} \;,
\end{aligned}
\end{equation}
where $\{\Delta \phi_k\}_{k=1}^3$ is measured at the gateway, as detailed in the next subsection.
Equation~\eqref{eqn_phase_shifts} involves three unknown parameters: the soil electrical permittivity $\epsilon$ and the signal direction $\vec{r_{\text{TX}}}$, defined by two angles~(elevation and azimuth). 
With three equations and three unknowns, we can solve for all of them.
This approach leverages the 3D geometry of the tetrahedral array, enabling three projections along diverse directions to support accurate phase shift measurements regardless of device placement or orientation.

\begin{figure*}[t]
    \subfigure[Tetrahedron antenna.]{
\includegraphics[width=.17\textwidth]{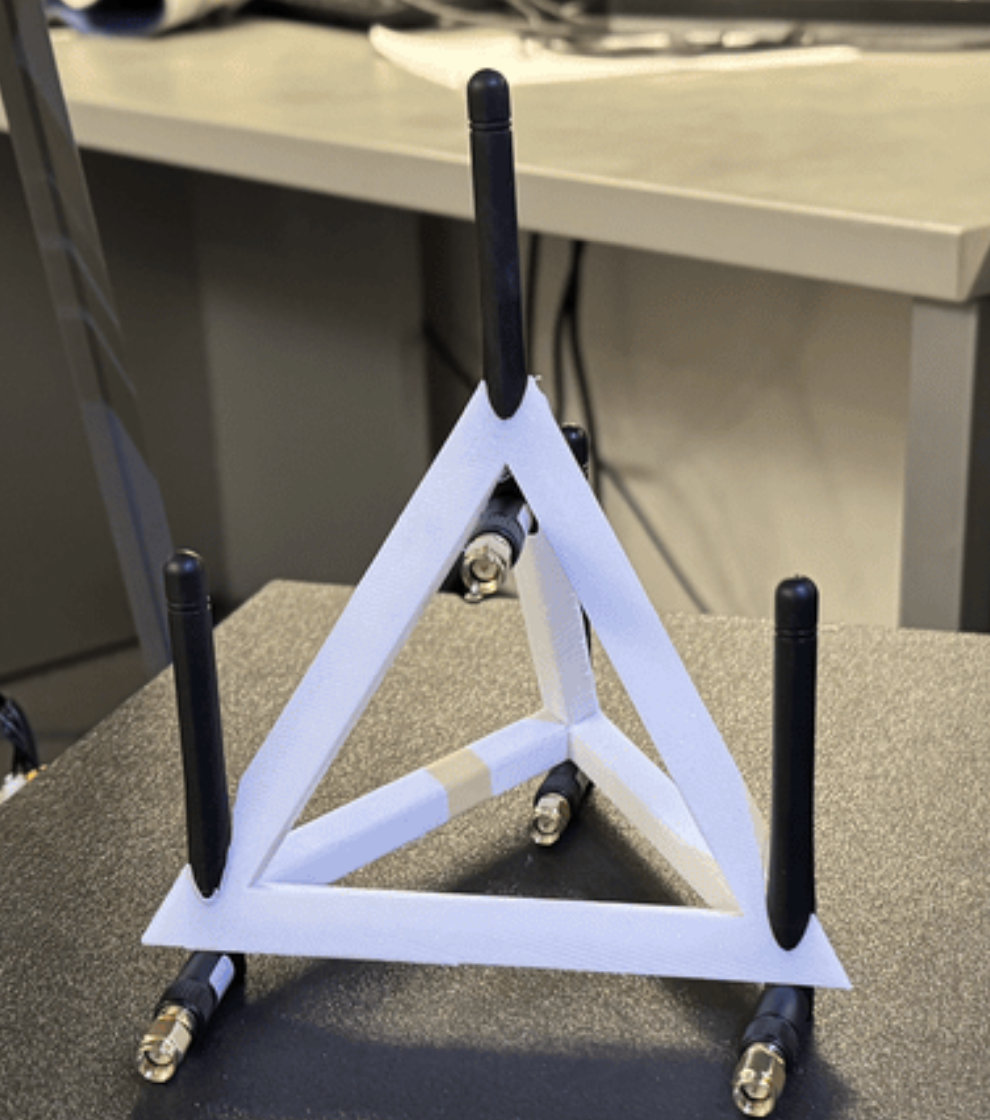}\label{fig_implementation_te}}
    \subfigure[LoRa transmitter.]{
\includegraphics[width=.26\textwidth]{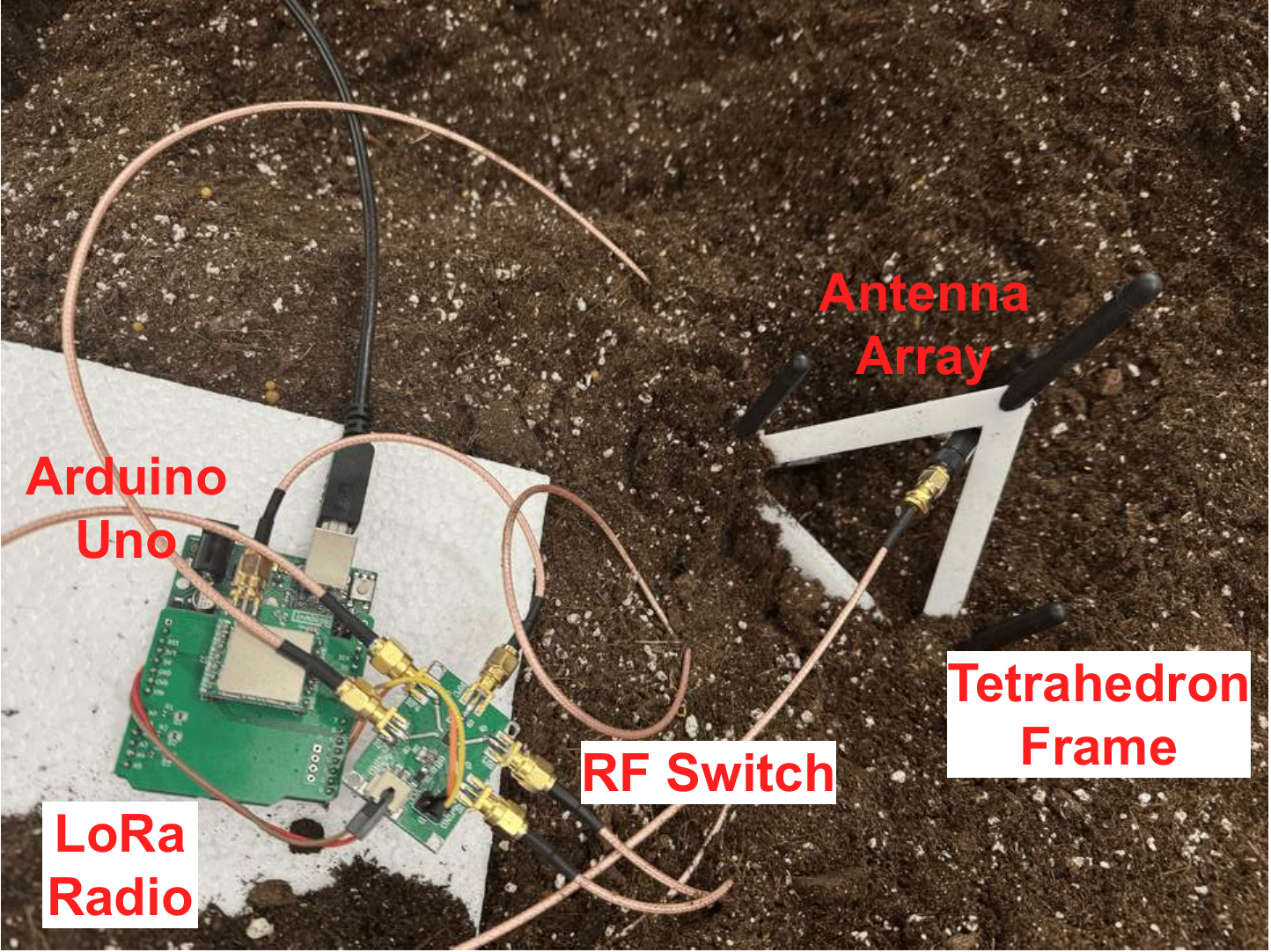}\label{fig_implementation_lora}}
    \subfigure[VNIR module.]{
\includegraphics[width=.26\textwidth]{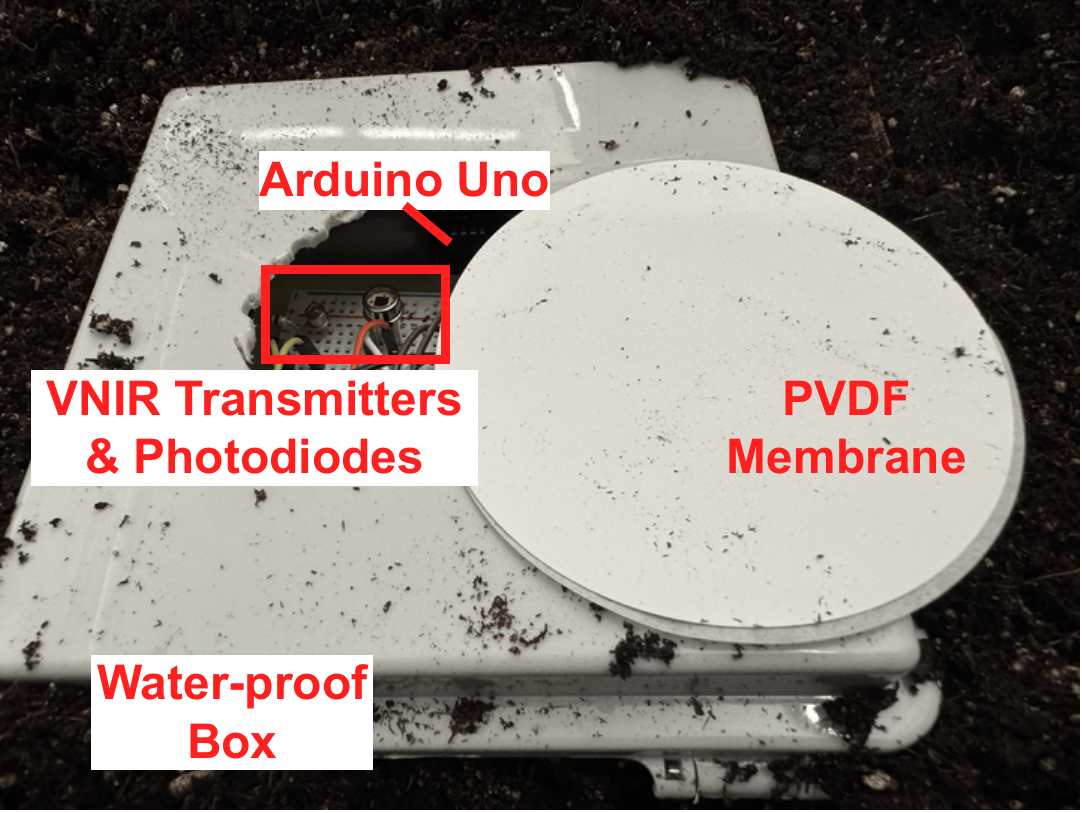}\label{fig_implementation_vnir}}
    \subfigure[In-field deployment.]{
\includegraphics[width=.26\textwidth]{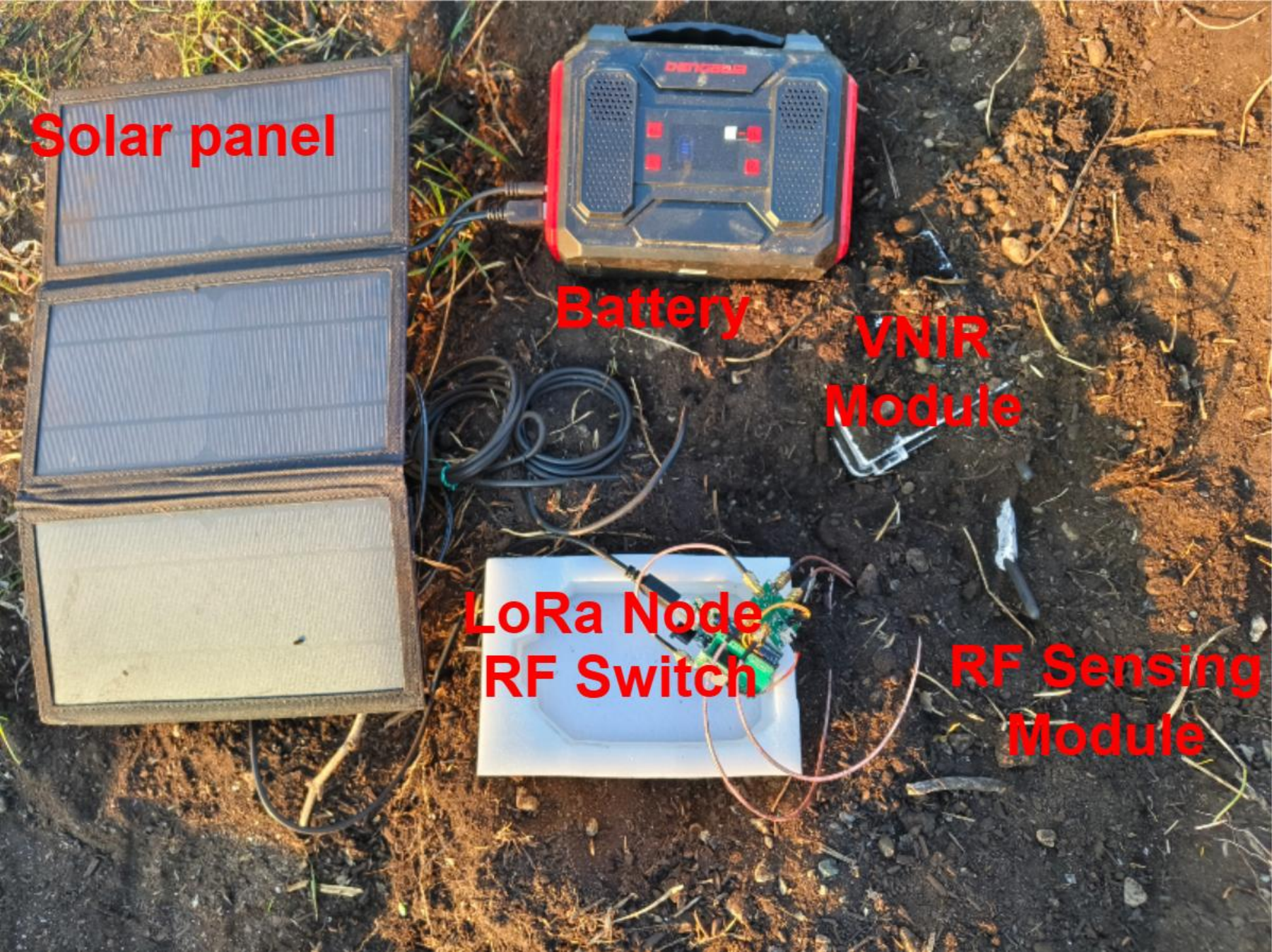}\label{fig_implementation_field}}
\caption{\ourSystem's LoRa and VNIR sensing hardware implementation, along with in-lab and in-field deployment.}
 \Description[]{}
    \label{fig_implementation}
\end{figure*}

\subsubsection{Antenna Switching Mechanism}

Antenna switching in a dual-antenna system is proposed to eliminate phase measurement errors for accurate phase shift $\Delta \phi$ measurement~\cite{chang2022sensor}. 
Specifically, when sending a LoRa packet, the preamble chirp is first transmitted via antenna 1, then switches to antenna 2. 
To extend this scheme to our four-antenna design for accurate phase shift measurement, we replace the single pole double throw~(SPDT) switch with a single pole four throw~(SPFT) switch. 
As depicted in Figure~\ref{fig_tera_switch}, during the transmission of a LoRa packet, the RF switch is initially connected to port 1, transmitting a portion of the preamble chirps through antenna 1. 
The switch then sequentially connects to ports~2, 3, and 4, with subsequent portions of the preamble transmitted via antennas 2, 3, and 4, respectively. 
Each switch transmits for a duration $t \in \left(T, 2T\right]$, where $T$ is the chirp duration.

At the gateway, we divide the received~\(\mathcal{IQ}\) samples from two consecutive preamble chirps to compute the chirp ratio.
If the antenna is switched between the two chirps, the chirp ratio yields the desired phase shift.
This process cancels out~CFO and phase offset errors~\cite{chang2022sensor}.
The rationale is described in~\cite{chang2022sensor}.
Processing consecutive chirp pairs yields three phase shifts~$\{\Delta \phi_k\}_{k=1}^3$, corresponding to antenna pairs between the origin antenna and the other three antennas.
The three phase shift measurements, together with the three unknown parameters, make Equation~\eqref{eqn_phase_shifts} solvable, enabling computation of soil permittivity~\(\epsilon\) regardless of device placement.
Section~\ref{sec_eval_robustness} shows that the tetrahedral design provides orientation-invariant permittivity and maintains stable component errors under yaw, pitch, and roll rotations as well as random throws.

\section{{Implementation}}\label{sec_implementation}

\begin{figure}
\centering
{\includegraphics[width=0.4\textwidth]{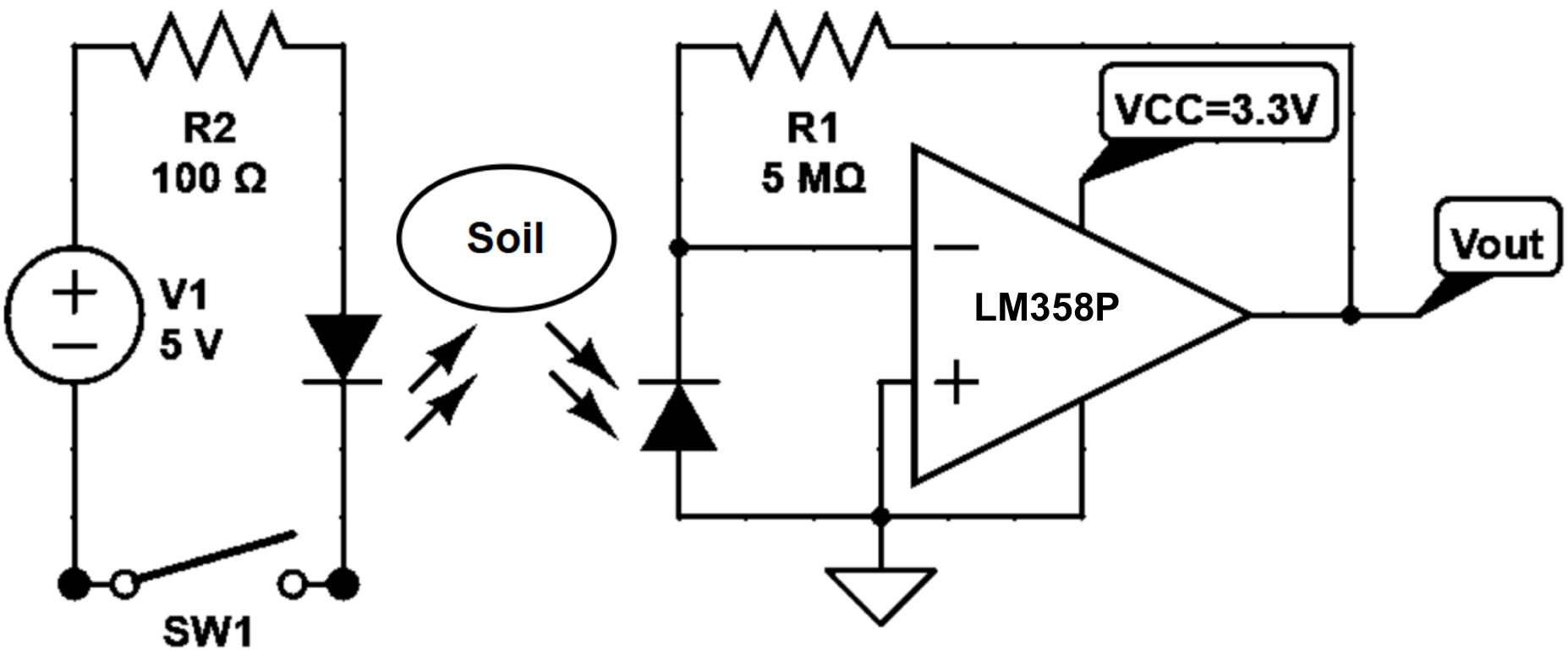}}
\vspace{2pt}
  \caption{Circuit topology of the VNIR sensing module.}
  \Description[]{}
    \label{fig_implement2}
\end{figure}

We build a \ourSystem prototype. It costs only \$44.8, which is significantly cheaper than a commercial soil meter (\$500)~\cite{Meter}.
As validated in~§\ref{sec_evaluation}, \ourSystem achieves lower sensing error than the commercial meter.

\textbf{LoRa Hardware.}
Figures~\ref{fig_implementation_te} and~\ref{fig_implementation_lora} illustrate the LoRa sensing node, which builts on an Arduino UNO host board with an~SX1276 LoRa Radio~(\$16.99)~\cite{kdd_marlp, mobisys_orchloc}, a 3D-printed tetrahedral antenna array~(\$6.5), and an HMC241 RF switch module~(\$9)~\cite{RFSWitch}.
The LoRa packet transmission is controlled via an open-source Arduino library, enhanced with custom timer modifications to ensure precise antenna switching during the packet's preamble transmission.

In the tetrahedral antenna array, the antenna spacing is set to 13.2~cm, ensuring phase differences remain within~$\left(-\pi, \pi\right)$ for the LoRa signal wavelength.  
3D printing of the tetrahedral frame costs less than \$0.2.  
The antenna switching is configured to occur every~1.5 chirp periods.  
Thus, switching through all antennas in the array requires less than 8 chirp periods, taking at most~32.768~ms~(with $SF = 9$ and $BW = 125~\text{kHz}$).  
This setting maintains compact hardware design and negligible time complexity.

\textbf{VNIR Hardware.} 
Figure~\ref{fig_implementation_vnir} shows VNIR sensing kit~(\$5.47), which includes seven Light Emitting Diodes (LEDs)~\cite{Bulb} operating at wavelengths of 460~nm, 620~nm, 1200~nm, 1300~nm, 1450~nm, 1550~nm, and 1650~nm, each selected for its sensitivity to specific soil components.
Controlled by an Arduino UNO host board, these LEDs emit light toward the soil, and the reflected light is captured by two photodiodes, \ie MTPD1346D-030 and KPDE030-H8-B~(\$5.9)~\cite{photodiodide}.
The photodiodes convert the reflected light into electrical signals, which are amplified by LM358P operational~(\$0.24) amplifiers to enhance signal strength and reduce noise.

Figure~\ref{fig_implement2} depicts the circuit topology: a 5V DC source~(V1) powers the photodiode through a 100~$\Omega$ resistor~(R2) and a switch~(SW1), while the amplifier~(LM358P), configured with a 5~M$\Omega$ feedback resistor~(R1) and powered by 3.3V~(VCC), amplifies the signal to produce~$V_\text{out}$.
This setup ensures high-sensitivity optical measurements of soil reflective properties.

\textbf{Membrane.}
As shown in Figure~\ref{fig_implementation_vnir}, a circular Polyvinylidene Difluoride~(PVDF) \cite{Membrane} membrane~(\$0.5) is centrally placed within the VNIR module, encapsulating soil samples to absorb moisture and diffuse trace elements onto its surface. 
This membrane, housed in a waterproof box along with an Arduino Uno, VNIR transmitters, and photodiodes, provides a stable interface for soil analysis and enhances light interaction with soil components.
The reflected light, rich in spectral information about the soil composition, is captured by photodiodes for further processing.

Finally, Figure~\ref{fig_implementation_field} shows the in-field deployment, where the VNIR module is integrated with a solar panel for power, a battery for energy storage, and LoRa sensing nodes for data transmission and monitoring in natural soil environments.

\textbf{Practicality.}
The LED–photodiode architecture mounts directly on a compact PCB and uses off-the-shelf drivers with an LM358-class amplifier, avoiding bulky gratings and alignment-sensitive spectrometers (Figure~\ref{fig_implement2}). 
Combined with the PVDF sampling interface and the solar-powered field node (Figure~\ref{fig_implementation_field}), the module remains robust and low-maintenance in outdoor deployments, supporting long-term operation on a small battery or solar supply.

\begin{figure}
\centering
{\includegraphics[width=0.3\textwidth]{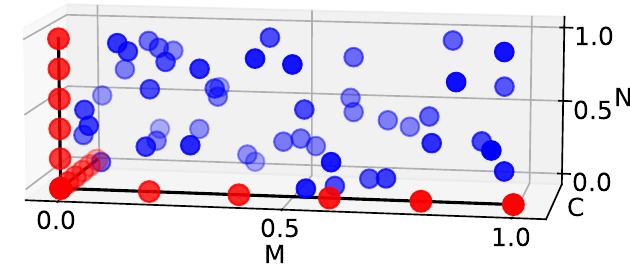}}
\caption{Illustration of the normalized values of three soil components~\(\{M, N, C\}\) in the training (red dots) and randomly distributed testing (blue dots) samples.}
  \Description[]{}
    \label{fig_vis_raw}
\end{figure}

\section{{Evaluation}}\label{sec_evaluation}

We compare \ourSystem with baseline systems using both in-lab manually formulated soil samples and in-field experiments.

\subsection{Experimental Settings}\label{sec_exp_setting}

We evaluate \ourSystem under in-lab conditions in~§\ref{sec_lab}, followed by an in-field study and robustness evaluation in~§\ref{sec_filed} and~§\ref{sec_eval_robustness}.
Next, §\ref{sec_ablation_study} presents ablation studies, while~§\ref{sec_eval_cl} compares \ourSystem with various contrastive learning algorithms. 
Finally, §\ref{sec_eval_overhead} analyzes the overhead.

\subsubsection{Datasets}
The training dataset consists solely of manually formulated soil samples, while the testing dataset includes both manually formulated and in-field soil samples.

\textbf{Training Dataset.}
We formulate 43 structurally labeled soil samples, as mentioned in §\ref{sec_otho}.
The example list is provided in Table~\ref{tab_training_data_small}.  
This dataset serves as the pre-training dataset for \ourSystem.
After pre-training, \ourSystem generalizes to different soil conditions without any retraining or fine-tuning.

\textbf{Test Dataset.}  
\textit{i)~For formulated testing soil samples,} each of the six components is randomly selected within typical agricultural value ranges: $M,C\in[0,50\%]$, $Al\in[0,10\%]$, and $NPK \in [0, 1\text{\textperthousand}]$. 
This differs from the structured training data, where only one component value is varied from the reference.
In total, we prepare~55 formulated testing samples, representative of diverse agricultural soils~\cite{patoine2022drivers,jian2020database}.

Figure~\ref{fig_vis_raw} illustrates the values of three soil components from the training and testing datasets in a 3D space, where red dots represent training samples and blue dots indicate testing samples. 
Each axis corresponds to one soil component. 
Training samples lie along the axes, reflecting controlled variations in which only one component is adjusted at a time. 
In contrast, testing samples are scattered throughout the space, representing combinations of multiple varying components. 
This demonstrates that the testing data goes beyond the training patterns, posing a stronger challenge for the model to generalize under realistic soil composition variability.

\textit{ii)~For in-field testing soil samples,} seven samples are collected from real agricultural fields using clean stainless steel tools.  
Furthermore, \ourSystem is deployed at the campus riverside for continuous monitoring over 12 days.  
These tests evaluate the generalizability of \ourSystem in real-world enviroments.

\begin{figure}
\centering
{\includegraphics[width=0.47\textwidth]{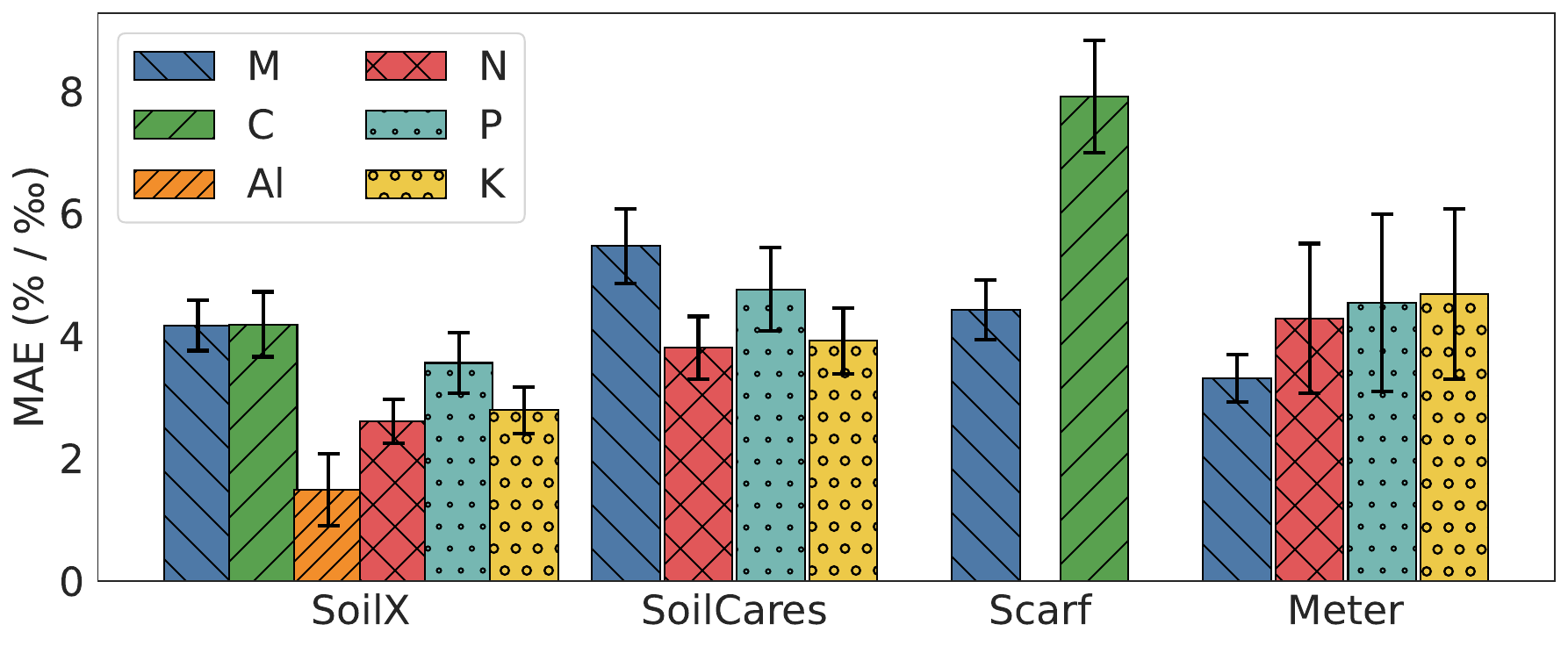}}
\caption{In-lab results: \ourSystem measures all six soil components, while others are limited.}
\Description[]{}
    \label{fig_eval_lab}
\end{figure}

\textbf{Ground Truth.}
For each formulated soil sample, we ensure that the defined component values~\(\left[ y_M, y_N, y_P, y_K, y_C, y_{\text{Al}} \right]\) reflect the actual composition of the prepared sample.
This is achieved by:
i)~vendor-supplied standard sand, silt loam, and clay are blended in calculated ratios to achieve the target \( y_{\text{Al}} \).
ii)~exact amounts of the other five components are added to match their defined values.
For example, high-purity fertilizers are added, including urea, phosphate, and potash containing 46\%~$N$, 46\%~$P$, and 60\%~$K$, respectively.
This ensures all training and testing samples have known and accurate six-component compositions.

For in-field samples, rather than relying on commercial sensors as ground truth, as in existing works~\cite{wang2024soilcares, chang2022sensor}, which remain insufficiently accurate, we sent soil samples to the Laboratory. 
The resulting measurements, in units such as~$g\cdot kg^{-1}$,~$g\cdot L^{-1}$, and~$ppm$, are then converted into percentages by calculating with conversion factors and density.
This provides trustworthy ground truth for five components~\(\left[ y_M, y_N, y_P, y_K, y_C \right]\).
Note that~\( y_{\text{Al}} \) is excluded, as multiple minerals~(\eg kaolinite, illite, montmorillonite) contain aluminum, making quantification difficult even in the laboratory.

\subsubsection{Performance Criteria}
For each soil component, we report the mean absolute error~(MAE) separately.
Since the value ranges for \(C\), \(M\), and \(Al\) are 0–50\%, while those for \(N\), \(P\), and \(K\) are 0–1\textperthousand, the MAEs are expressed in \% for \(C\), \(M\), and \(Al\), and in \textperthousand{} for \(NPK\).

Aluminosilicate texture~$Al$ is analyzed as a key confounding factor influencing both sensing and learning. 
A quantitative metric for~$Al$ is reported only when reliable ground truth exists; otherwise, it is treated as a latent variable used to decouple cross-component interference and is not considered a validated outcome.

\textbf{Reporting Policy.}
We distinguish between (i) laboratory blends with known texture ground truth and (ii) in-field deployments without laboratory analysis. 
In case~(i), $Al$ is reported as both a validated metric and auxiliary target; in case~(ii), $Al$ accuracy is omitted, and the variable is used only internally within the model. 
All tables and figures follow this policy.

\begin{figure}
\centering
{\includegraphics[width=0.45\textwidth]{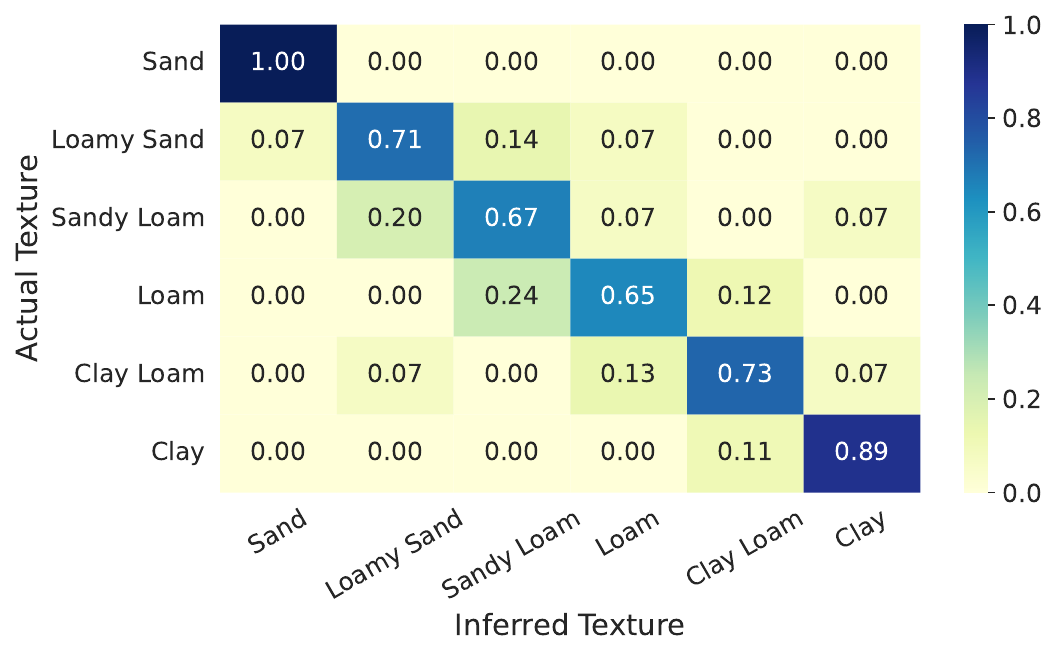}}
  \caption{Soil texture classification with $Al$ output.}
  \Description[]{}
    \label{fig_eval_texture}
\end{figure}

\begin{figure*}[t]
 \subfigure[Campus riverside soil.]{
\includegraphics[width=.18\textwidth]{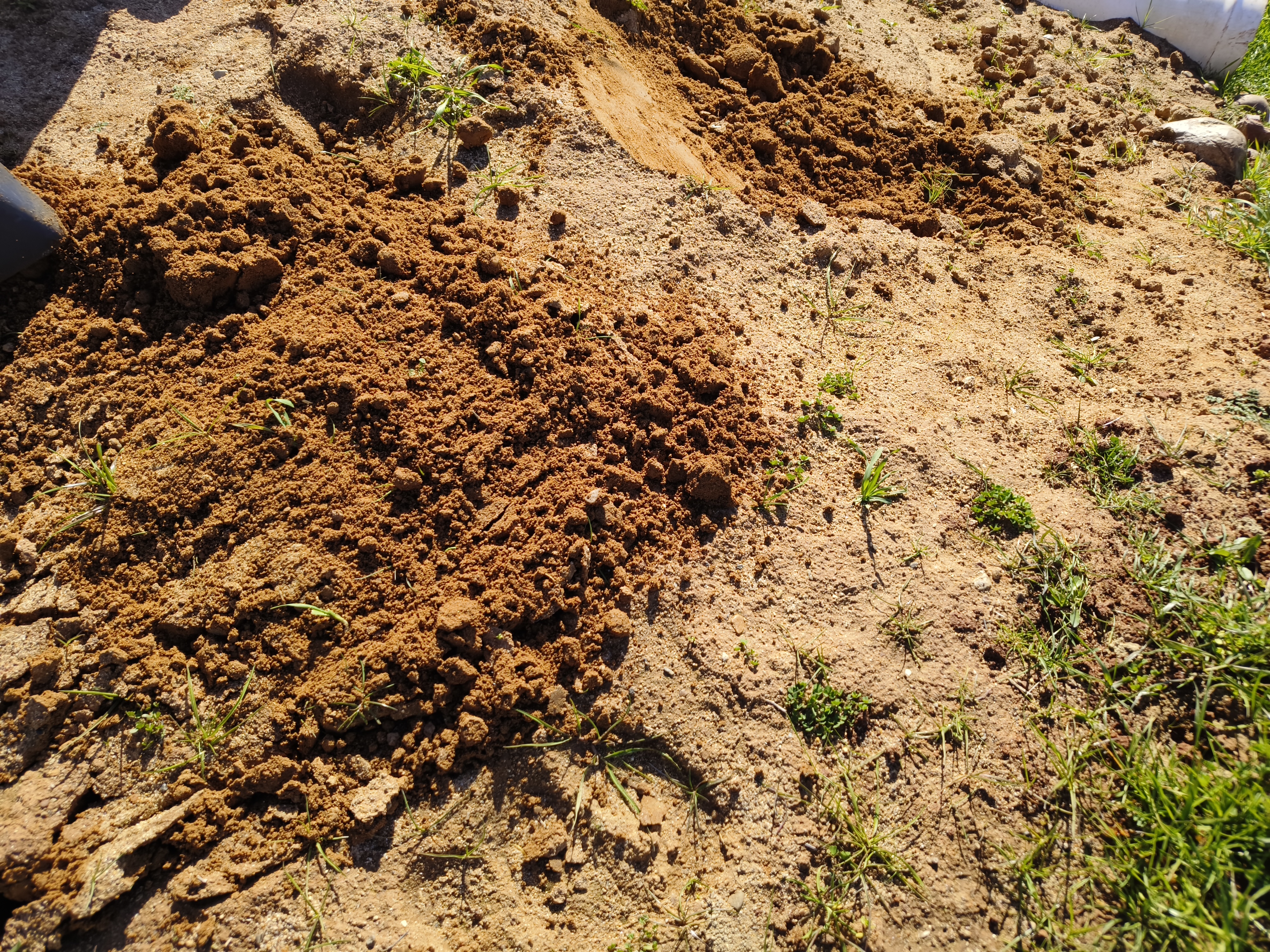}
\label{fig_soil_res}}
    \subfigure[Walnut soil.]{
\includegraphics[width=.18\textwidth]{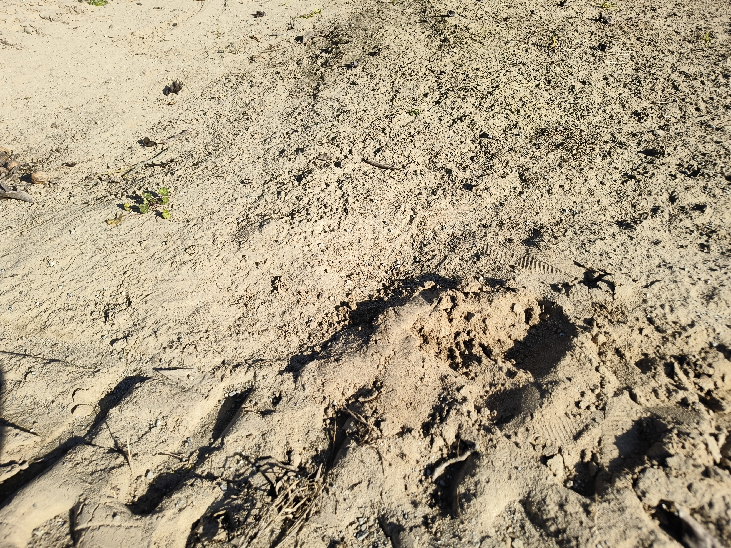}}
 \subfigure[Mound soil.]{
\includegraphics[width=.18\textwidth]{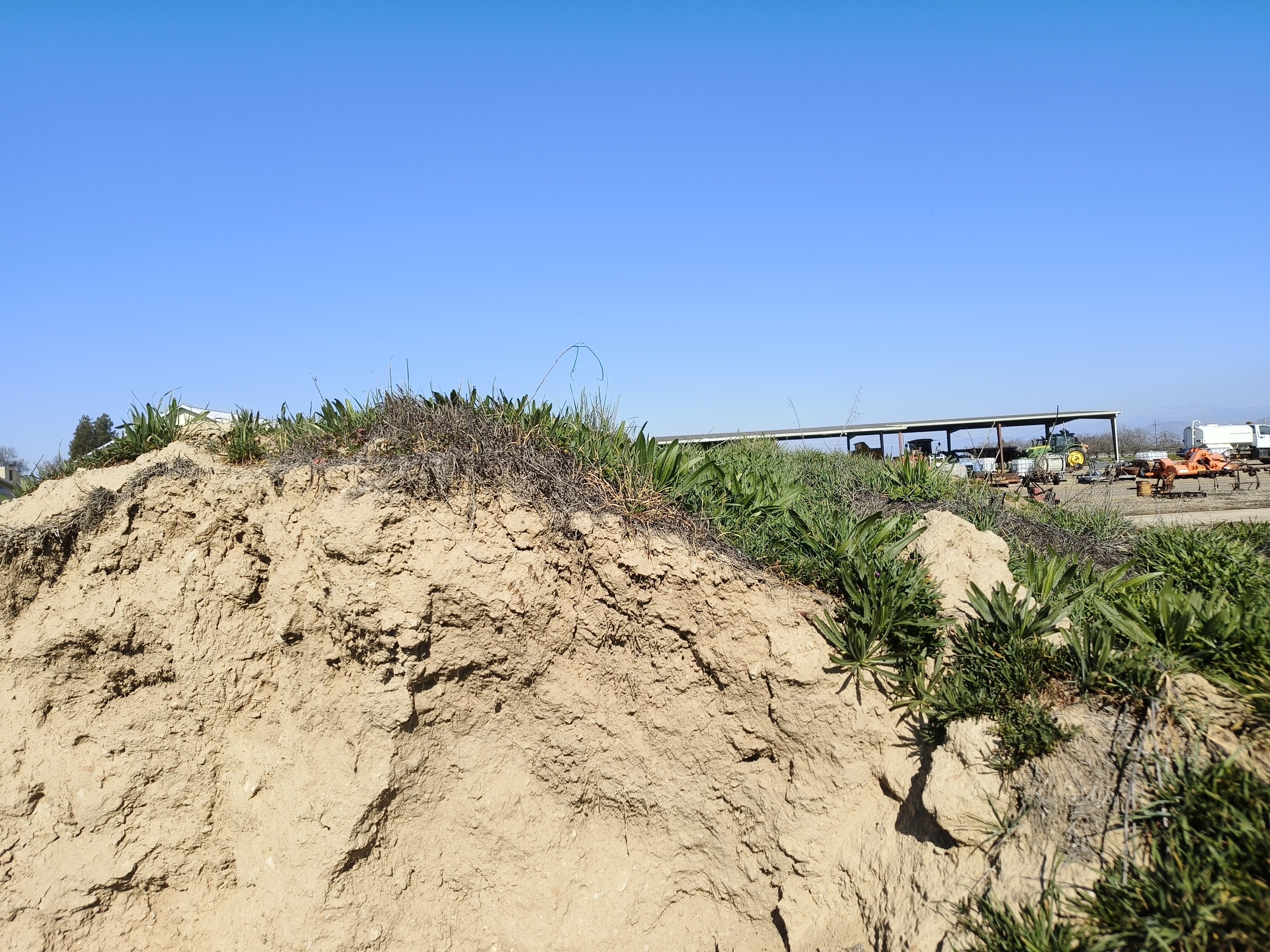}}
 \subfigure[Alfalfa soil.]{
\includegraphics[width=.18\textwidth]{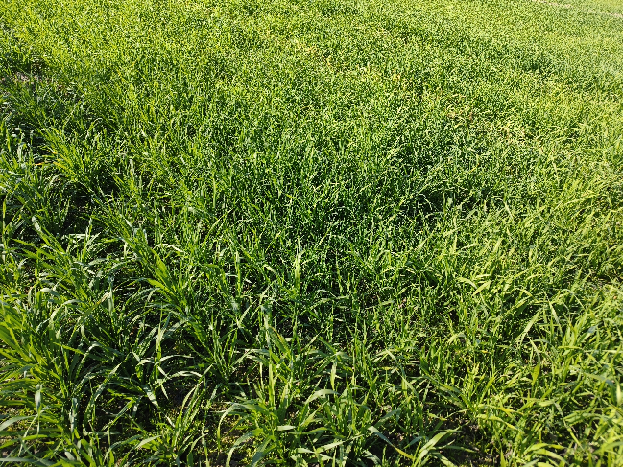}}
 \subfigure[Almond soil.]{
 \label{fig_soil_almond}
\includegraphics[width=.18\textwidth]{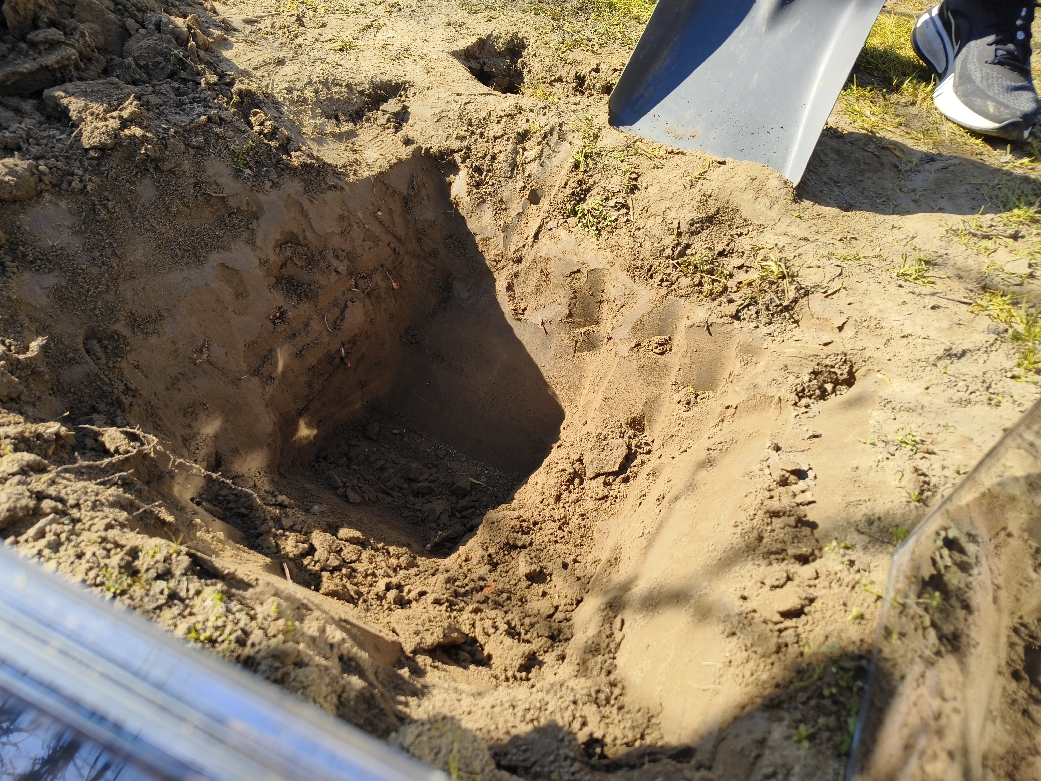}}
\caption{Five in-field scenarios covering diverse soil types: riverside, walnut, mound, alfalfa, and almond soils.}
 \Description[]{On-site scenarios.}
    \label{fig_area}
\end{figure*}

\subsubsection{Baselines}
The following three baselines are included:

\begin{itemize}[label=\textbullet, leftmargin=1em, topsep=1pt]

\item SoilCares~\cite{wang2024soilcares}: Utilizes a dual-antenna LoRa node and a VNIR sensing module to sequentially estimate moisture and then~$NPK$.
Specifically, we reproduce the hardware and apply a random forest model trained on the same dataset as \ourSystem.

\item Scarf~\cite{ding2024cost}: 
Combines CSI data and optical signals sequentially, \ie first measuring~$M$, then~$C$. 
We use LoRa hardware to extract CSI, as LoRa is widely adopted in agricultural environments~\cite{ren2024demeter}. 
The model takes grayscale images along with the CSI as input for~\(\{M, C\}\) estimation during both training and test. 
The performance is close to the values reported  in the original article.

\item Meter~\cite{Meter}: A commercial soil testing device for measuring soil moisture and~$NPK$ component values.

\end{itemize}

Note that both SoilCares~\cite{wang2024soilcares} and Scarf~\cite{ding2024cost} baselines use the same training and testing data as \ourSystem to ensure a fair comparison.

\subsection{In-lab Experiments}\label{sec_lab}

\begin{figure}
\centering
{\includegraphics[width=0.46\textwidth]{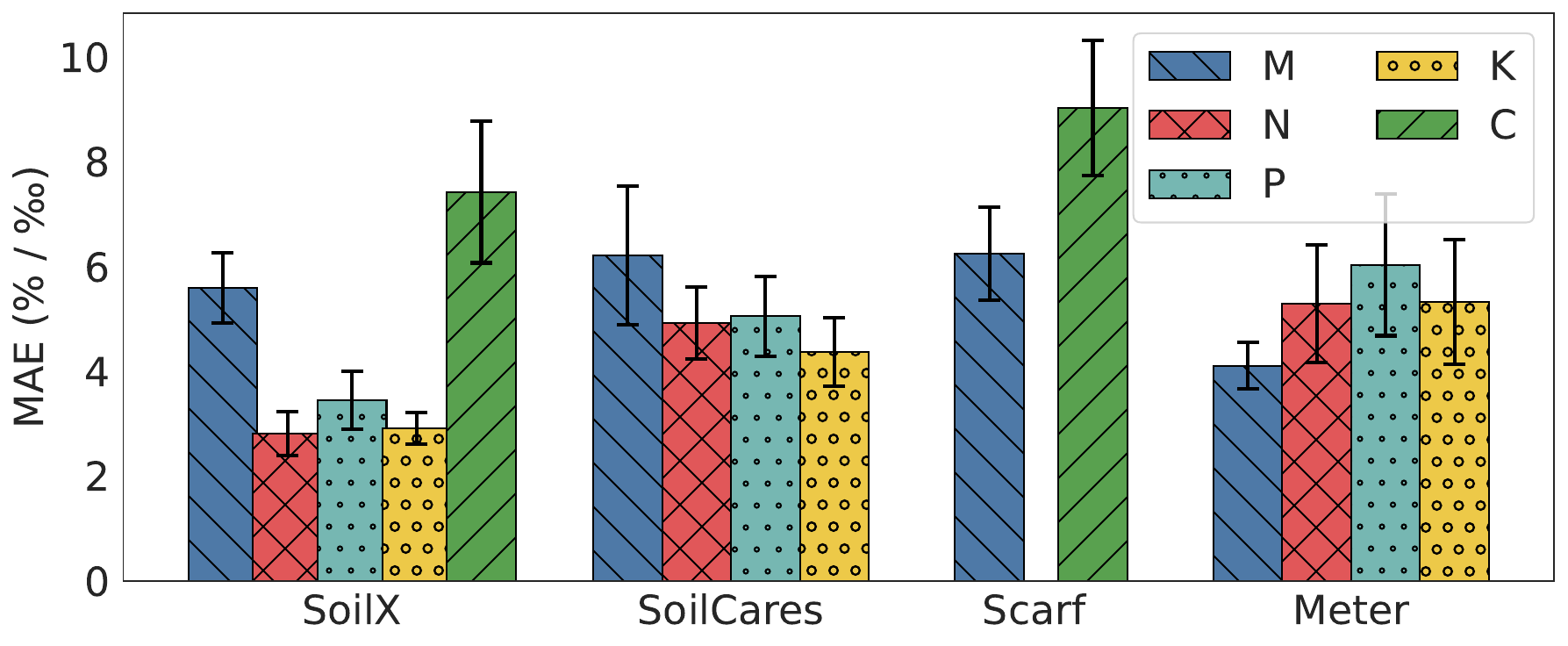}}
\caption{Performance under in-field experiments.}
\Description[]{}
    \label{fig_eval_field}
\end{figure}

\subsubsection{Overall Performance}\label{sec_performance}

We evaluate \ourSystem's performance in a controlled lab environment to assess its error in quantifying all six components simultaneously. 
We randomly prepare soil samples as test data, featuring varying compositions that span the same component ranges as the training dataset. 
Each sample is analyzed using~\ourSystem, along with three baseline models for comparison.

In Figure~\ref{fig_eval_lab}, the in-lab results demonstrate that \ourSystem achieves an average MAE of 4.17\% for~$M$, representing a 23.8\% reduction over SoilCares' random forest regression and a 5.9\% reduction over Scarf~\cite{ding2024cost}. 
\ourSystem also achieves accuracies of 4.19\% for organic carbon and 2.61\textperthousand, 3.56\textperthousand, and 2.79\textperthousand{} for~$N$,~$P$, and~$K$, respectively. 
Compared to SoilCares, the improvements for~$NPK$ range from 25.2\% to 31.5\%. 
For organic carbon, \ourSystem achieves a 47.0\% reduction in MAE compared to Scarf. 
For validation, a commercial soil meter priced at \$449 achieves 3.31\% for moisture, 4.29\textperthousand{} for~$N$, 4.54\textperthousand{} for~$P$, and 4.68\textperthousand{} for~$K$, demonstrating that \ourSystem delivers comparable or better performance at a fraction of the cost.

The MAE for aluminum is 1.49\%. 
However, aluminum content is not a direct indicator for treatment guidance; rather, it is used to infer soil texture. 
To this end, we assess classification performance based on the inferred aluminum values. 
The confusion matrix in Figure~\ref{fig_eval_texture} shows an overall classification accuracy of 76.92\%. 
Sand achieves 100\% accuracy, followed by clay at 88.89\%, while intermediate textures such as clay loam, loamy sand, sandy loam, and loam yield lower accuracies of 73.33\%, 71.43\%, 66.67\%, and 64.71\%, respectively. 
Most misclassifications occur between adjacent categories, reflecting the gradual nature of soil texture transitions. 
For instance, loamy sand is partially misclassified as sandy loam, and clay loam is misclassified as loam or clay. 
These results indicate that \ourSystem effectively captures soil texture characteristics and demonstrates strong potential for soil texture identification.

\subsection{In-Field Study}\label{sec_filed}

Figure~\ref{fig_implementation_field} presents the in-field deployment of \ourSystem. 
We select five geographically distributed field-test sites, as illustrated in Figure~\ref{fig_area}. 
Specifically, at the campus riverside location~(Figure~\ref{fig_soil_res}), the deployment spans 12 days, covering varying temperatures, wind speeds, and a total of 14 hours of rainfall. 
In the almond orchards~(Figure~\ref{fig_soil_almond}), we conduct experiments at different soil depths. 
Ground truth measurements are obtained through physical sampling followed by laboratory analysis.

Note that in-field ground truth for~$Al$ is not available; thus,~$Al$ results are not reported.
Even lab-based X-ray fluorescence~(XRF) spectrometers cannot fully capture the~\(Al\) proportion across a soil sample, as multiple minerals~(\eg kaolinite, illite, montmorillonite) contain~\(Al\), making comprehensive quantification difficult.

\begin{figure}
\centering
{\includegraphics[width=0.45\textwidth]{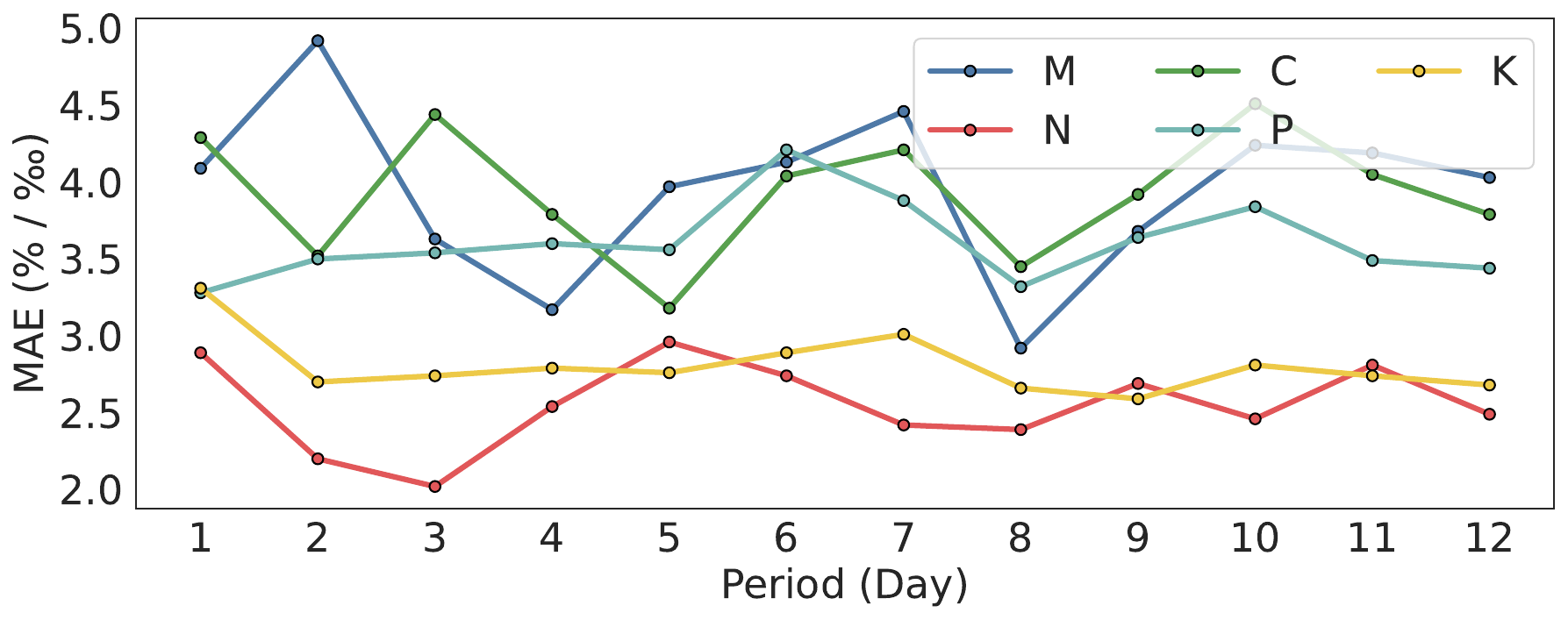}}
\caption{Performance of \ourSystem at location~(a), the campus riverside site, across different days.}
\Description[]{}
\label{fig_eval_field_temporl}
\end{figure}

\textbf{Experimental Setup and Deployment Geometry.}
Each field deployment is conducted outdoors at five representative sites~(Figure~\ref{fig_area}). 
At each site, we excavate a shallow trench and install a complete \ourSystem node—including the VNIR module, tetrahedral LoRa array, MCU, and power unit—inside a weather-sealed enclosure~(Figure~\ref{fig_implementation_field}). 
The trench is backfilled with native soil so that the sensing surfaces and antennas are buried, while the PVDF membrane window remains flush with the ground to interface with pore water and protect the electronics from direct wetting. 
Ground truth is collected through periodic nearby sampling followed by laboratory assays. 
Deployments operate continuously for multiple days under natural weather conditions, ensuring results reflect realistic field performance rather than controlled benchtop tests.

\begin{figure}
\centering
{\includegraphics[width=0.45\textwidth]{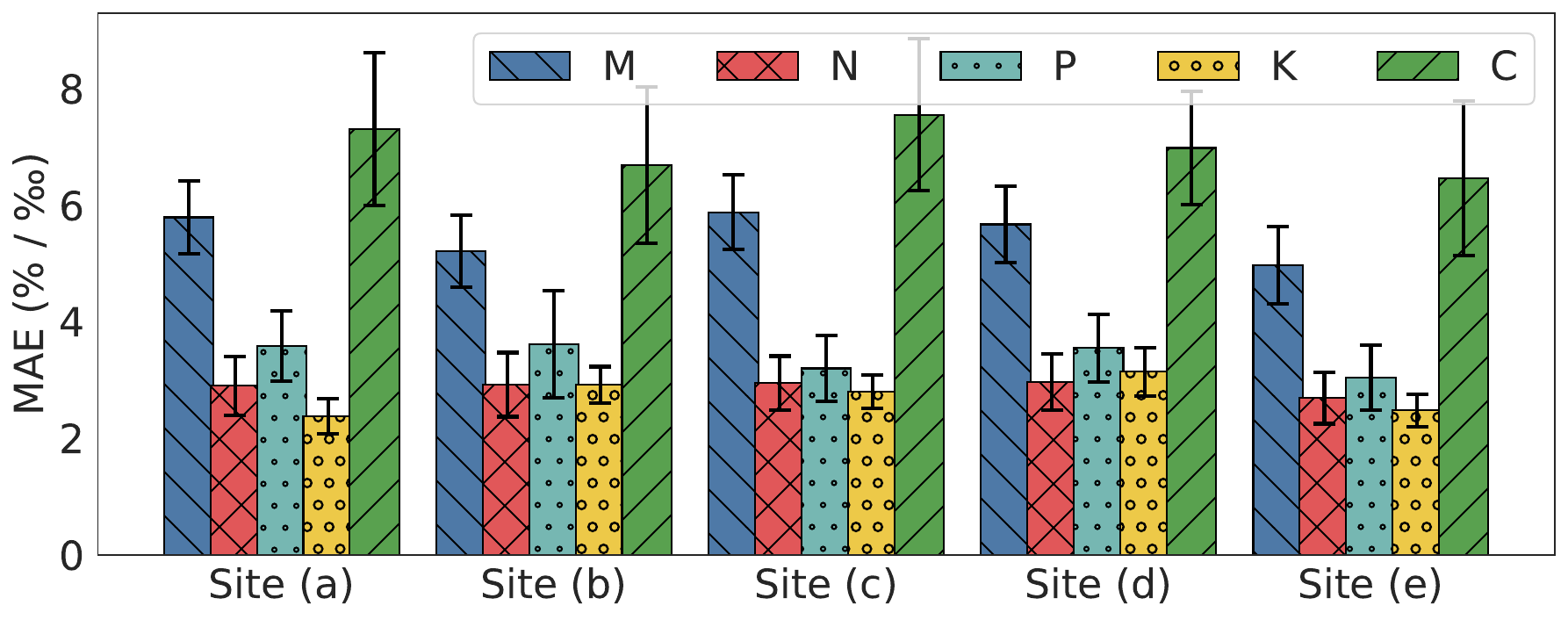}}
\caption{Performance of \ourSystem across five diverse sites during the same experimental period.}
\Description[]{}
    \label{fig_eval_field_sptial}
\end{figure}

\textbf{Overall.}
We present the overall performance, then analyze \ourSystem's effectiveness across spatial and temporal dimensions. 
As demonstrated in Figure~\ref{fig_eval_field}, \ourSystem achieves MAEs of 5.60\% for moisture, 2.81\textperthousand{} for~$N$, 3.45\textperthousand{} for~$P$, and 2.91\textperthousand{} for~$K$. 
In contrast, SoilCares yields higher estimation errors: 6.21\% for moisture, 4.92\textperthousand{} for~$N$, 5.05\textperthousand{} for~$P$, and 4.37\textperthousand{} for~$K$. 
Scarf reports MAEs of 6.25\% for moisture and 9.03\% for carbon, both significantly higher than those of \ourSystem. 
These results highlight \ourSystem's generalization performance in in-field component quantification, validating both its system design and the effectiveness of our training sample preparation.

\textbf{Benefits of \ourSystem over SoilCares and Scarf.}
Across five field sites (Figure~\ref{fig_eval_field}), \ourSystem achieves lower mean absolute error on $\{M,N,P,K\}$ and accurately estimates $C$. 
It requires no site-specific calibration or antenna alignment, unlike SoilCares, which assumes stable texture, or Scarf, which relies on precise optical setup. 
With its orientation-insensitive LoRa array and 3CL framework that suppresses cross-component interference, \ourSystem offers robust, scalable, and maintenance-free operation in real agricultural conditions.

\textbf{Temporal and Spatial Performance.}
Figures~\ref{fig_eval_field_temporl} and~\ref{fig_eval_field_sptial} further illustrate \ourSystem's performance across different times and locations.  
Figure~\ref{fig_eval_field_temporl} indicates that \ourSystem achieves an average~$M$ estimation error of~3.95\% with a standard error of~0.52\%, demonstrating stable performance despite environmental variations such as temperature, wind, and rainfall.
Figure~\ref{fig_eval_field_sptial} presents spatial performance across five test sites (Figure~\ref{fig_area}).
The MAEs for~$M$ range from~4.97\% to~5.88\%, for~$N$ from~2.69‰ to~2.97‰, for~$P$ from~3.04‰ to~3.61‰, for~$K$ from~2.38‰ to~3.14‰, and for~$C$ from~6.46\% to~7.55\%.
This narrow variation across all components reflects \ourSystem's consistent accuracy in diverse geographical conditions, likely influenced by local factors such as soil type, topography, and microclimate.
Together, these results confirm that \ourSystem delivers reliable and accurate soil sensing under real-world conditions.

\begin{figure}[t]
\centering
    \subfigure[Ground truth.]{
    \label{fig_eval_depth_gt}
\includegraphics[width=.50\linewidth]{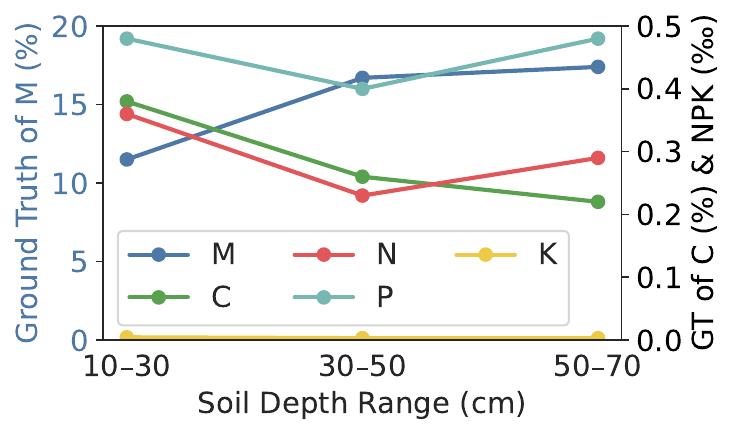}}
        \hfill
    \subfigure[MAE.]{
    \label{fig_eval_depth_mae}
\includegraphics[width=.46\linewidth]{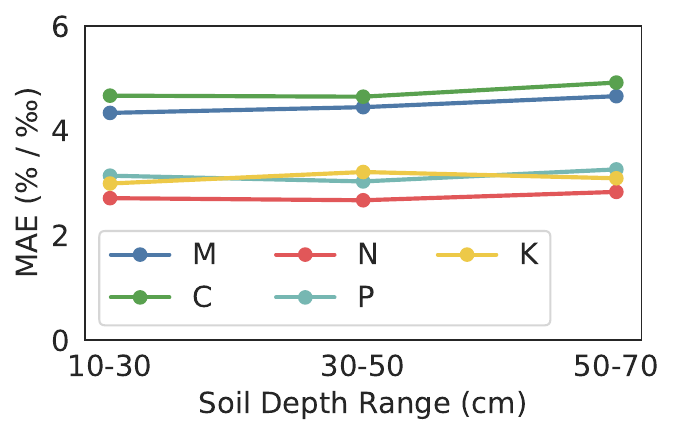}}
\caption{Performance at different device depths.}
\Description[]{}
\label{fig_eval_depth}
\end{figure}

\begin{figure}[t]
\centering
{\includegraphics[width=0.45\textwidth]{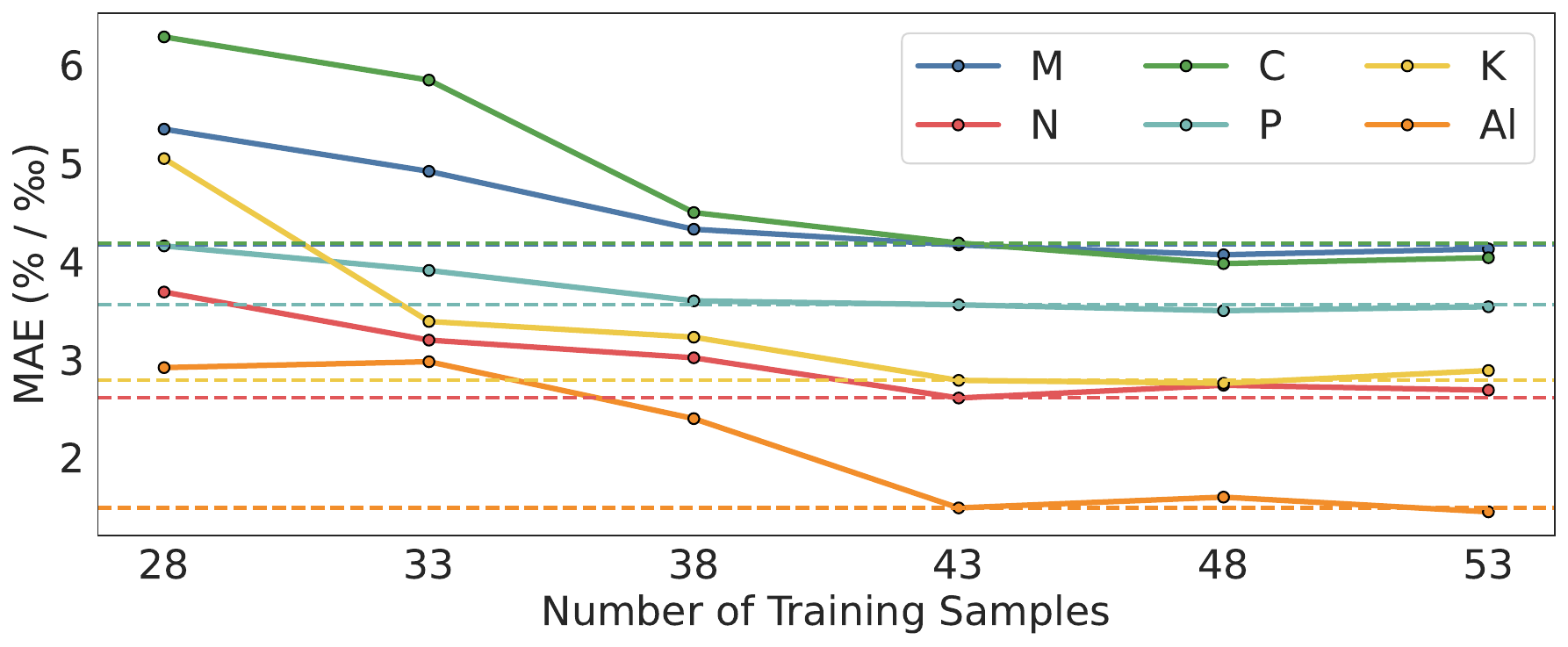}}
\caption{Impact of training data amount on performance.}
\Description[]{}
    \label{fig_eval_converge}
\end{figure}

\subsubsection{Impact of Depth Change}
\label{sec_exp_depth}

Soil properties vary slightly with depth due to differences in physical and chemical composition. 
To evaluate the impact of depth on soil content and inference performance, we deploy \ourSystem at three depths~(10–30~cm, 30–50~cm, and 50–70~cm) at the same location. 
Simultaneously, soil samples are collected from these depths and analyzed in the lab to obtain ground truth measurements. 
The results in Figure~\ref{fig_eval_depth_gt} show that moisture content increases with depth, while organic carbon decreases, and macronutrient concentrations remain relatively stable.
As shown in Figure~\ref{fig_eval_depth_mae}, \ourSystem maintains stable performance across all depths, with only slight increases in error: 2.1\% and 3.6\%.
This minor degradation is primarily due to the accumulation effect of~\(\epsilon\).
For example, while the ground truth is based on samples from the 50–70~cm layer, the LoRa signal traverses the entire 0–70~cm soil volume.

\subsubsection{Impact of Sensing Range}\label{sec_sensing_range}

To evaluate the sensing range limit, we assess the variation in~\(\epsilon\) by changing the distance between the LoRa sensing node and the gateway. 
Starting at 0~m, where the gateway is positioned directly above the antennas, we incrementally increase the distance in 20~m steps and collect sensing signals, \ie the sensing vector~\(\mathbf{x}_{s}\). 
The measured~\(\epsilon\) values at 0~m, 20~m, 40~m, and 60~m are 22.3, 21.6, 21.4, and 22.1, respectively. 
Taking 0~m as the reference, the deviations range from 0.9\% to 4.0\%, which are negligible. 
At 80~m,~\(\epsilon\) cannot be calculated due to undetectable phase changes. 
Thus, the reliable sensing range in open space is 60~m, consistent with Chang~\etl~\cite{chang2022sensor}. 
This confirms that the tetrahedral array enhances~E2U without compromising the sensing range.

\subsection{Robustness Study}
\label{sec_eval_robustness}

\subsubsection{Impact of Training Data Amount}

We do not include cross-validation, as our method intrinsically requires structurally labeled data. 
Instead, we evaluate model convergence and robustness by varying the number of customized training samples. 
Additional samples are randomly selected from the full numerical range of each soil component to maintain coverage diversity.

\begin{figure}[t]
\centering
{\includegraphics[width=0.46\textwidth]{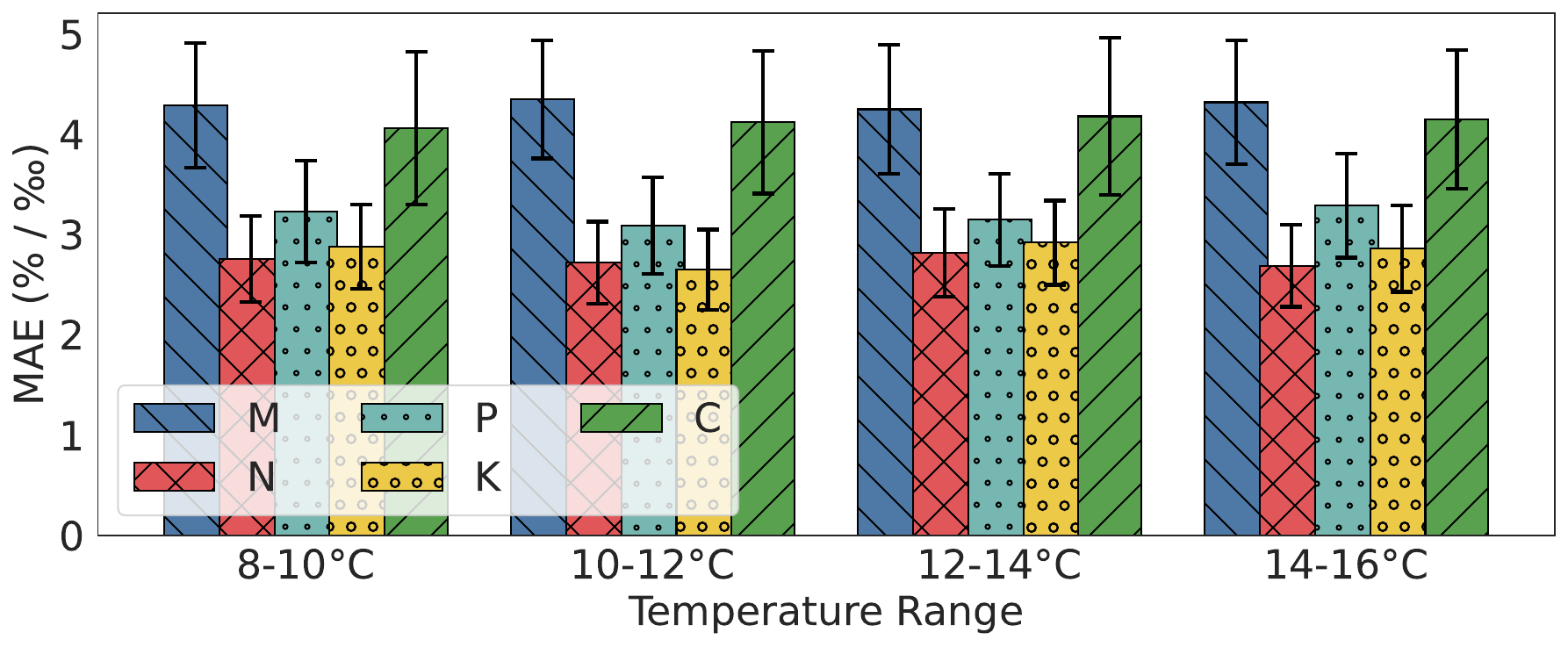}}
  \caption{Performance under different temperatures.}
  \Description[]{}
    \label{fig_eval_temp}
\end{figure}

\begin{figure}[t]
\centering
{\includegraphics[width=0.46\textwidth]{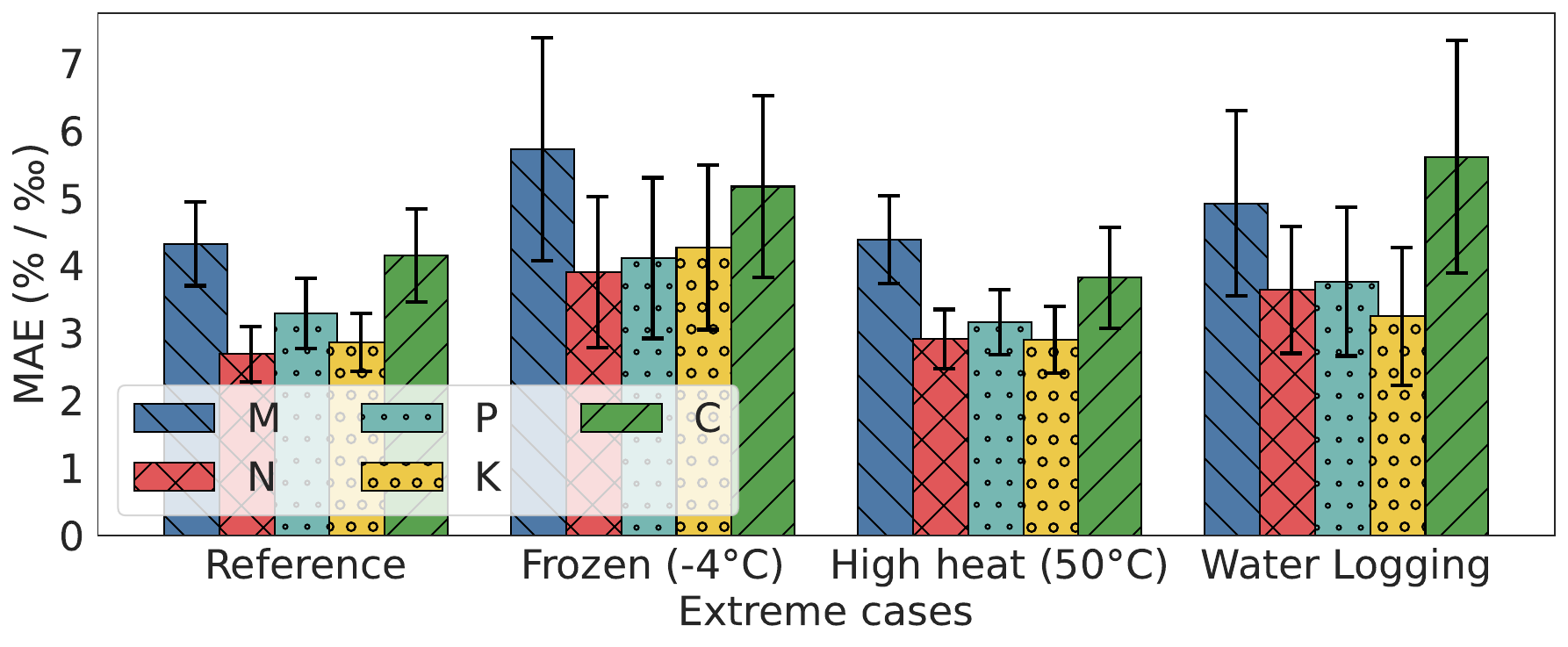}}
  \caption{Performance under extreme weather cases.}
  \Description[]{}
    \label{fig_eval_extreme}
\end{figure}

As shown in Figure~\ref{fig_eval_converge}, performance generally improves as more training samples are included. 
For example, the MAE for~$N$ decreases from 3.69\textperthousand{} to 2.61\textperthousand{} as the number of training samples increases from 28 to 53.
Similar downward trends are observed for other components such as~$K$,~$P$, and~$M$, indicating better generalization with increased sample diversity. 
In contrast, some components like$Al$ and~$C$ exhibit relatively stable performance regardless of data size, suggesting that their spectral or permittivity signatures may already be well captured with fewer samples.
Overall, these results demonstrate that \ourSystem, equipped with compound contrastive loss and structurally labeled data, achieves stable and accurate convergence even with a relatively small training set, and can further benefit from modest increases in training sample size.

\subsubsection{Impact of Weather Conditions}

We evaluate \ourSystem's performance under different weather scenarios by categorizing the experiments into \textit{Normal} and \textit{Extreme} conditions.

\textbf{i)~Normal.}  
In typical field scenarios with soil temperatures ranging from 8$^\circ$C to 16$^\circ$C, precipitation and wind exert negligible influence on inference error~(<0.1\%). 
While the pre-training soil samples are collected at 19.8$^\circ$C, test measurements are categorized into four temperature intervals: 8--10$^\circ$C, 10--12$^\circ$C, 12--14$^\circ$C, and 14--16$^\circ$C.
Figure~\ref{fig_eval_temp} indicates that temperature variations cause only minor shifts in soil dielectric properties, leading to small changes in prediction errors: 
$M$~(4.25\%--4.35\%), 
$N$~(2.69\textperthousand--2.82\textperthousand), 
$P$~(3.09\textperthousand--3.29\textperthousand), 
and 
$K$~(2.65\textperthousand--2.92\textperthousand).
These results confirm \ourSystem's stable performance under normal environmental conditions.

\textbf{ii)~Extreme.}  
We further evaluate robustness under extreme conditions by subjecting samples to freezing~($-4^\circ$C), high heat~(50$^\circ$C), and waterlogging.  
Freezing increases light reflection due to icing, causing the absorption at 620~nm and 460~nm to decrease by 19.7\% to 90.4\%, while other wavelengths remain relatively stable, with an average deviation of 2.2\%.  
Under freezing, the~$NPK$ error increases by up to 62.7\%~(Figure~\ref{fig_eval_extreme}), whereas high heat shows minimal effect.  
As these conditions exceed typical sensing limits, further studies are needed to better understand and mitigate their impact.

\subsubsection{Impact of Antenna Placement}
\begin{figure}
\centering
{\includegraphics[width=0.46\textwidth]{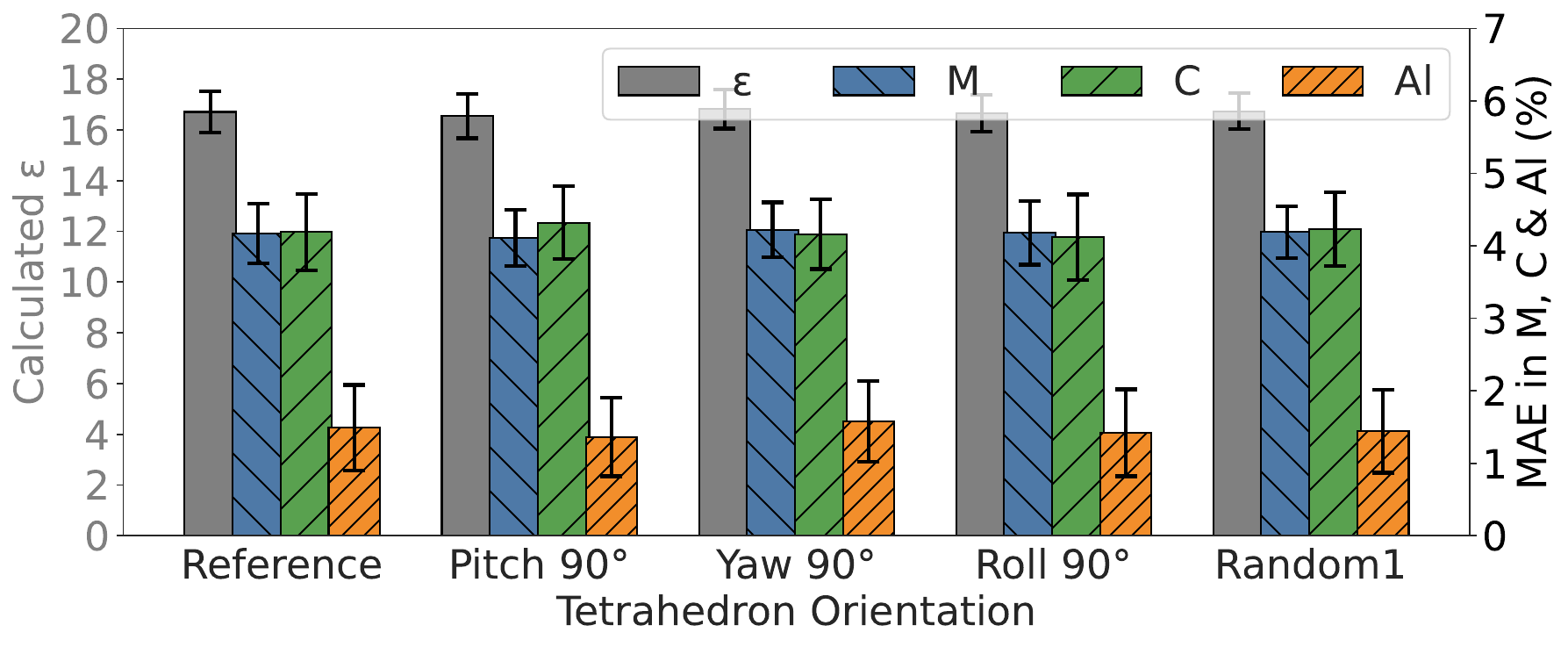}}
  \caption{Performance under different orientation.}
  \Description[]{}
    \label{fig_eval_orien}
\end{figure}

Equation~(11) for the tetrahedral array is independent of the array's orientation, theoretically enabling \ourSystem to function regardless of placement. We validate this by rotating the device by $90^\circ$ along the yaw, pitch, and roll axes, and by simulating real-world use through three ``random-throw'' tests that leave the array in arbitrary orientations.
Across all cases, including the regular orientation, the average dielectric permittivity is essentially unchanged at $\epsilon=16.69\pm0.14$ (see Fig.~\ref{fig_eval_orien}). Soil-moisture, carbon, and aluminum estimates are similarly stable, with mean absolute errors of $M: 4.17\pm0.06\%$, $C: 4.22\pm0.11\%$, and $Al: 1.49\pm0.13\%$, respectively. Measurements of $N$, $P$, and $K$ are omitted here because they depend on the VNIR module rather than LoRa placement.
These results confirm that antenna placement has minimal impact: the four-antenna tetrahedral design provides geometric diversity and infers permittivity without orientation assumptions, eliminating the need for alignment during deployment and reducing placement overhead.

\begin{figure}
\centering
{\includegraphics[width=0.46\textwidth]{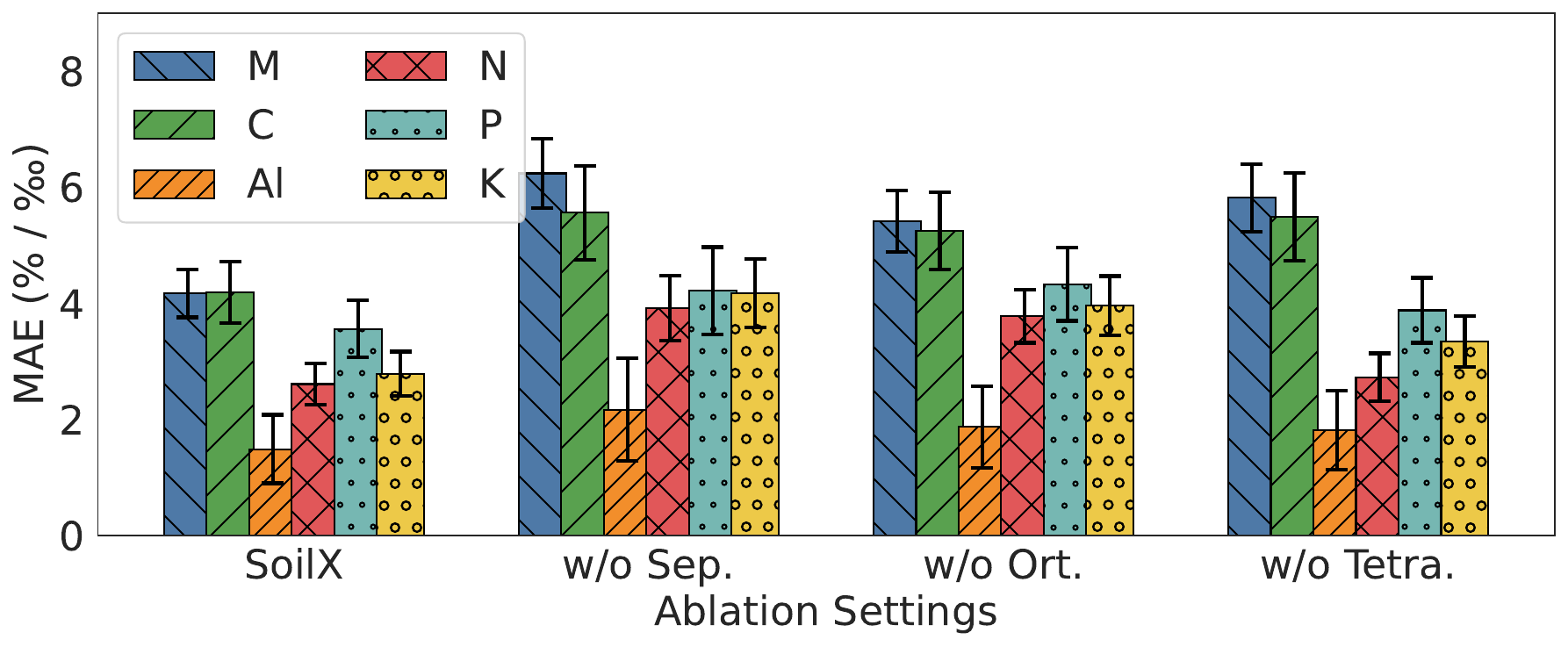}}
  \caption{Performance for ablated versions of \ourSystem.}
  \Description[]{}
    \label{fig_eval_ablation}
\end{figure}

\subsubsection{Impact of VNIR Plane Inclination.}\label{sec_vnir_plane}

Since the VNIR sensing kit is separate from the antenna tetrahedron, we assess its performance under varying inclinations.  
To simulate potential placement errors, the kit is inclined at $0^\circ$, $15^\circ$, $30^\circ$, and $45^\circ$ in a silt loam soil.  
The voltage readings across all frequency channels remain highly stable, within a fluctuation range of 0.03~V, indicating negligible impact.  
This robustness is attributed to the stable PVDF membrane, which ensures reliable performance despite placement deviations.

\subsection{Ablation Study}\label{sec_ablation_study}

We evaluate the contributions of \ourSystem’s components through an ablation study by removing or modifying specific elements: Separation Loss~(Sep.), Orthogonality Regularizer~(Ort.), and the tetrahedron antenna array~(Tetra.).  
The \textit{w/o Sep.} version uses only the Orthogonality Regularizer as the loss function, while \textit{w/o Ort.} employs only the Separation Loss.  
For \textit{w/o Tetra.}, the tetrahedron antenna is replaced with a standard dual-antenna setup that utilizes only a single phase difference.  
All configurations are trained and evaluated using in-lab formulated soil samples, with randomized antenna array placements to simulate practical deployment conditions.

As illustrated in Figure~\ref{fig_eval_ablation}, removing the Separation Loss results in the largest increase in MAE, with an average rise of 49.8\%, underscoring its central role in disentangling complex sensing signals.  
Eliminating the Orthogonality Regularizer increases the error by 33.4\% on average, highlighting its contribution to maintaining independence among soil component directions in the latent space.  
Replacing the tetrahedron antenna with a dual-antenna setup degrades soil permittivity estimation, increasing the errors for~$M$,~$C$, and~$Al$, while causing only minor degradation for~$N$,~$P$, and~$K$, which primarily depend on VNIR sensing.  
These results confirm that all components—Separation Loss, Orthogonality Regularizer, and tetrahedral antenna design—play complementary and critical roles in ensuring robust and accurate multi-component soil sensing.

\begin{figure}
\centering
{\includegraphics[width=0.46\textwidth]{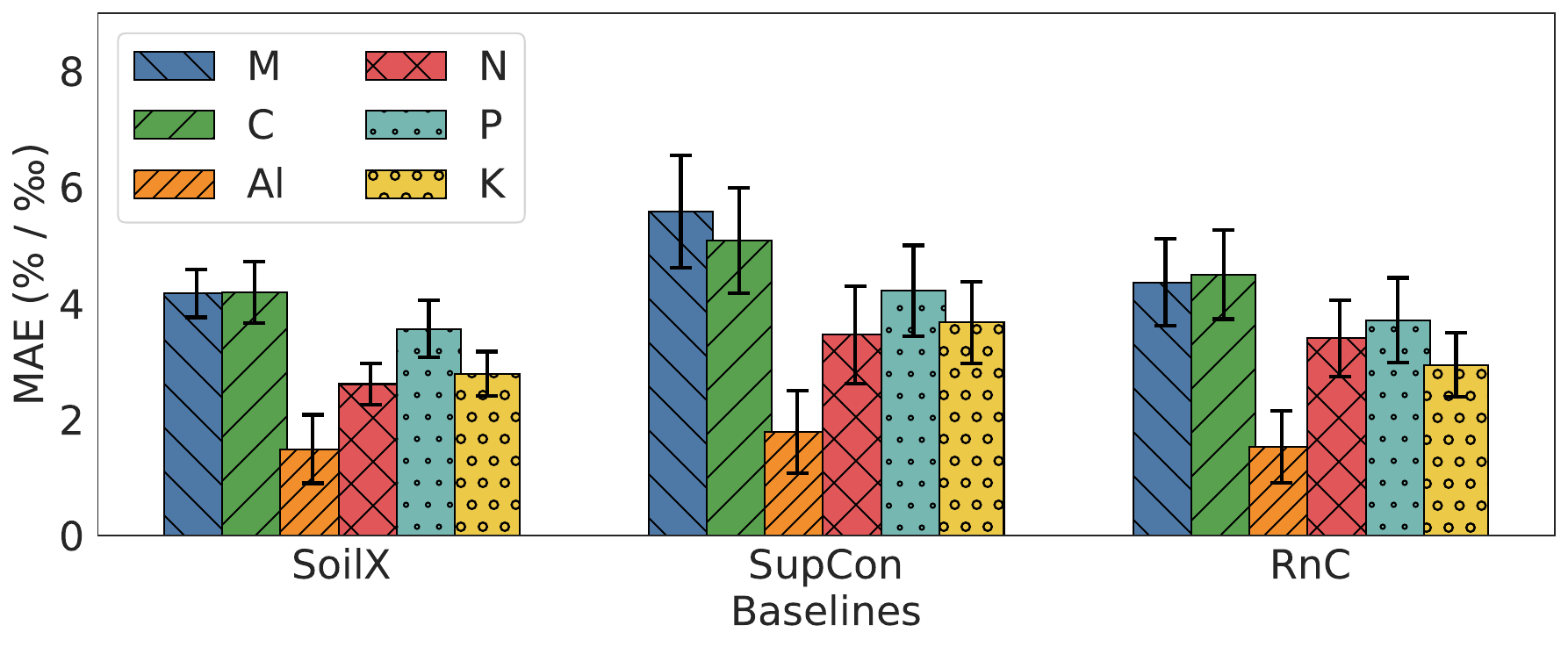}}
  \caption{Contrastive learning baseline comparison.}
  \Description[]{}
    \label{fig_eval_cl}
\end{figure}

\subsection{Different Contrastive Learning Schemes}\label{sec_eval_cl}

To assess the effectiveness of our 3CL framework, we compare it with canonical supervised CL~(SupCon)~\cite{khosla2020supervised} and regression-based CL~(RnC)~\cite{zha2023rank}, both trained and tested on the same data as \ourSystem. 
As shown in Figure~\ref{fig_eval_cl}, 3CL achieves consistently lower mean absolute error, improving by 15 to 25\% over SupCon and by 4 to 23\% over RnC across all six soil components: moisture~($M$), nutrients~($N,P,K$), organic carbon~($C$), and aluminosilicate texture~($Al$). 
These results show that the separation and orthogonality mechanisms in 3CL effectively handle multi-variable regression and limited data, offering clear advantages over classical contrastive learning approaches.

\subsection{Overhead Study}\label{sec_eval_overhead}

\textbf{Computation.}
\ourSystem is designed for lightweight training and inference. 
The encoder contains two 512-unit hidden layers and one 6-unit output layer, requiring about $5.38\times10^5$~FLOPs per inference, which completes in roughly 60~ms on a 16~MHz microcontroller. 
Pre-training on an NVIDIA GeForce RTX~3080~Ti takes only 176~s, and the calibration-free design eliminates the need for site-specific fine-tuning, enabling rapid adoption and scalable deployment.

\textbf{Energy.}
Each node consumes 38~mW during active sensing and communication and only 0.5~mW in standby mode, which is lower than the 0.5--3~W power draw typical of industrial-grade soil meters. 
This low consumption allows continuous sensing without frequent recharging or maintenance. 
Powered by a 1000~mAh, 5~V battery that stores roughly 18{,}000~J of energy, the system sustains operation for more than ten days under normal duty cycles. 
When paired with a 21~W solar panel and housed in a waterproof enclosure, \ourSystem supports uninterrupted, autonomous field operation across extended periods and varying weather conditions, enabling scalable long-term deployment in remote agricultural environments.

\section{{Discussion}}

\textbf{Why Narrow-Band LEDs Instead of Broadband Sweep?}
For a field-deployable, battery-powered platform, sweeping the full VNIR spectrum with a broadband source and spectrometer is possible but impractical due to high optical and power costs, complex alignment, and frequent calibration needs, all of which reduce reliability in outdoor environments. 
\ourSystem instead uses a compact array of narrow-band LEDs centered at informative wavelengths, coupled with low-cost photodiodes, to capture the essential spectral signatures of soil chemistry. 
This design minimizes energy consumption, avoids redundant bands, and enables fast, robust measurements suitable for long-term autonomous deployment.

Guided by controlled measurements and prior soil-spectroscopy studies, we select discrete wavelengths most sensitive to the six key components: moisture ($\sim$1450\,nm), organic carbon ($\sim$1650\,nm), nitrogen ($\sim$1200\,nm), phosphorus ($\sim$620\,nm), and potassium ($\sim$460\,nm), along with two auxiliary bands (1300\,nm and 1550\,nm) that capture cross-component interactions and texture cues. 
This targeted design maintains diagnostic sensitivity while minimizing redundancy, optical complexity, and energy consumption per measurement.

\textbf{PVDF Interface as an Optical Interface.}
The PVDF membrane functions as a stable optical interface rather than a chemical adsorbent. 
In natural soils, aluminum and silicon remain bound within aluminosilicate minerals and do not diffuse into the membrane. 
Upon soil contact, the PVDF film rapidly forms a thin, consistent water layer that contains dissolved nutrients ($NPK$) and organic carbon ($C$). 
Light scattering and absorption through this interfacial layer encode the spectral features necessary for VNIR-based sensing of soil chemistry and texture, providing a reproducible optical path while protecting the underlying electronics.

\textbf{Optical Response Time of the PVDF Interface.}
Since the measurement relies on optical path length and spectral absorption rather than diffusion or mass transfer into the polymer, no extended equilibration period is required. 
Relevant ions and dissolved species reach the PVDF interface within minutes, matching the natural equilibration timescale of soil moisture and enabling rapid, repeatable measurements in the field.

\textbf{PVDF Benefits for Accuracy and Repeatability.}
The PVDF interface minimizes surface roughness effects and local shadowing from exposed soil grains, producing a more uniform illumination and collection geometry. 
This stabilization enhances the signal-to-noise ratio for $NPK$ and $C$ estimation while maintaining sensitivity to permittivity and texture variations, leading to more accurate and repeatable measurements in field conditions.

\textbf{PVDF Trade-offs and Limitations.}
The PVDF membrane introduces minor optical attenuation and requires periodic cleaning to maintain transparency. 
In strongly acidic soils or in the presence of chelating agents, trace $\mathrm{Al}^{3+}$ or $\mathrm{Si(OH)}_{4}$ may be slightly mobilized, though the sensing method does not depend on their adsorption. 
Regular wiping or occasional replacement of the film preserves measurement accuracy during extended field operation.

\textbf{Without the PVDF.}
Without the PVDF membrane, VNIR light interacts directly with bare soil, causing highly variable reflections dominated by abundant minerals such as Si and Al, which mask the weaker spectral signatures of $NPK$. 
Direct soil contact also exposes the photodiodes and circuitry to moisture and abrasion, leading to rapid signal degradation and reduced device longevity.

\section{{Conclusion}}\label{sec_conclusion}

This paper presents \ourSystem, a calibration-free soil sensing system that quantifies six major soil components via LoRa and VNIR sensing modalities. 
\ourSystem introduces a Contrastive Cross-Component Learning~(3CL) framework to disentangle cross-component interference for accurate multi-component measurement. 
Its tetrahedron-based LoRa antenna placement design ensures robust permittivity estimation.
Extensive evaluations demonstrate \ourSystem's effectiveness.

\begin{acks}
This publication was prepared with the support of a financial assistance award approved by the Economic Development Administration, Farms Food Future, and an UC Merced Fall 2023 Climate Action Seed Competition Grant. Xinyu Zhang was supported in part by an Ericsson Endowed Chair fund, and by the National Science Foundation through grant CNS-2312715.  
Any opinions, findings, and conclusions expressed in this material are those of the authors and do not necessarily reflect the views of the funding agencies.
\end{acks}

\bibliographystyle{ACM-Reference-Format}

\end{document}